\documentclass{article}
\usepackage[letterpaper, left=1in, right=1in, top=1in, bottom=1in]{geometry}
\usepackage{natbib}
\bibpunct{(}{)}{;}{a}{,}{,}
\usepackage[colorlinks,citecolor=blue!70!black,linkcolor=blue!70!black, urlcolor=blue!70!black,breaklinks=true]{hyperref}
\usepackage{parskip}
\usepackage[dvipsnames, table]{xcolor}
\definecolor{lightblue}{RGB}{220,235,250}
\definecolor{sotacolor}{RGB}{198,239,206}

\usepackage{microtype}
\usepackage{booktabs} 
\usepackage{amsmath}
\usepackage{dsfont} 
\usepackage{amssymb}
\usepackage{xspace}
\usepackage{url}            
\usepackage{algorithm}
\usepackage[noend]{algorithmic}
\usepackage{amsthm}
\usepackage{listings}
\usepackage{bm}
\usepackage{bbm}
\usepackage{enumitem}
\usepackage{nicefrac}       
\usepackage{mathrsfs} 
\usepackage{inconsolata}
\usepackage{wrapfig}
\usepackage{graphicx}
\usepackage{bigints}
\usepackage{transparent}
\usepackage{comment}
\usepackage[nameinlink,capitalize]{cleveref}
\makeatletter
\AddToHook{cmd/appendix/before}{
\def\cref@section@alias{appendix}
\def\cref@subsection@alias{appendix}
\def\cref@subsubsection@alias{appendix}
}
\makeatother
\usepackage{thmtools}
\usepackage{thm-restate}
\usepackage{nicematrix}
\usepackage{arydshln}
\usepackage{fancyhdr}
\usepackage{mathtools}
\usepackage[scaled=.88]{helvet}
\usepackage{breakcites}
\usepackage{multirow}
\usepackage{wasysym}
\usepackage{array}
\usepackage{tabularx}
\usepackage{svg}
\usepackage{subcaption}
\usepackage{enumitem}
\usepackage{siunitx}

\DeclareMathOperator{\unif}{{unif}}

\renewcommand{\epsilon}{\varepsilon}

\newtheoremstyle{spaced}
  {6pt}   %
  {0pt}   %
  {\itshape} %
  {}       %
  {\bfseries} %
  {.}      %
  {0.5em}  %
  {}
\theoremstyle{spaced}

\newcommand{\algcommentlight}[1]{\textcolor{blue!70!black}{\transparent{0.5}\small{\texttt{\textbf{//\hspace{2pt}#1}}}}}

\DeclarePairedDelimiter{\abs}{\lvert}{\rvert} %

\DeclarePairedDelimiter{\crl}{\{}{\}}

\DeclarePairedDelimiterX{\infdiv}[2]{(}{)}{%
  #1\;\delimsize\|\;#2%
}

\newcommand{\wh}[1]{\widehat{#1}}
\newcommand{\wb}[1]{\widebar{#1}}

\def\ddefloop#1{\ifx\ddefloop#1\else\ddef{#1}\expandafter\ddefloop\fi}
\def\ddef#1{\expandafter\def\csname bb#1\endcsname{\ensuremath{\mathbb{#1}}}}
\ddefloop ABCDEFGHIJKLMNOPQRSTUVWXYZ\ddefloop
\def\ddefloop#1{\ifx\ddefloop#1\else\ddef{#1}\expandafter\ddefloop\fi}
\def\ddef#1{\expandafter\def\csname b#1\endcsname{\ensuremath{\mathbf{#1}}}}
\ddefloop ABCDEFGHIJKLMNOPQRSTUVWXYZ\ddefloop
\def\ddef#1{\expandafter\def\csname sf#1\endcsname{\ensuremath{\mathsf{#1}}}}
\ddefloop ABCDEFGHIJKLMNOPQRSTUVWXYZ\ddefloop
\def\ddef#1{\expandafter\def\csname c#1\endcsname{\ensuremath{\mathcal{#1}}}}
\ddefloop ABCDEFGHIJKLMNOPQRSTUVWXYZ\ddefloop
\def\ddef#1{\expandafter\def\csname h#1\endcsname{\ensuremath{\widehat{#1}}}}
\ddefloop ABCDEFGHIJKLMNOPQRSTUVWXYZ\ddefloop
\def\ddef#1{\expandafter\def\csname hc#1\endcsname{\ensuremath{\widehat{\mathcal{#1}}}}}
\ddefloop ABCDEFGHIJKLMNOPQRSTUVWXYZ\ddefloop
\def\ddef#1{\expandafter\def\csname t#1\endcsname{\ensuremath{\widetilde{#1}}}}
\ddefloop ABCDEFGHIJKLMNOPQRSTUVWXYZ\ddefloop
\def\ddef#1{\expandafter\def\csname tc#1\endcsname{\ensuremath{\widetilde{\mathcal{#1}}}}}
\ddefloop ABCDEFGHIJKLMNOPQRSTUVWXYZ\ddefloop
\def\ddefloop#1{\ifx\ddefloop#1\else\ddef{#1}\expandafter\ddefloop\fi}
\def\ddef#1{\expandafter\def\csname scr#1\endcsname{\ensuremath{\mathscr{#1}}}}
\ddefloop ABCDEFGHIJKLMNOPQRSTUVWXYZ\ddefloop

\let\oldparagraph\paragraph
\renewcommand{\paragraph}[1]{\oldparagraph{#1}}

\renewcommand{\epsilon}{\varepsilon}

\newcommand{\ldef}{\vcentcolon=}

\renewcommand{\bigm}[1]{%
  \ifcsname fenced@\string#1\endcsname
    \expandafter\@firstoftwo
  \else
    \expandafter\@secondoftwo
  \fi
  {\expandafter\amsmath@bigm\csname fenced@\string#1\endcsname}%
  {\amsmath@bigm#1}%
}

\newcommand{\DeclareFence}[2]{\@namedef{fenced@\string#1}{#2}}
\makeatother

\makeatletter
\let\save@mathaccent\mathaccent
\newcommand*\if@single[3]{%
  \setbox0\hbox{${\mathaccent"0362{#1}}^H$}%
  \setbox2\hbox{${\mathaccent"0362{\kern0pt#1}}^H$}%
  \ifdim\ht0=\ht2 #3\else #2\fi
  }
\newcommand*\rel@kern[1]{\kern#1\dimexpr\macc@kerna}
\newcommand*\widebar[1]{\@ifnextchar^{{\wide@bar{#1}{0}}}{\wide@bar{#1}{1}}}
\newcommand*\wide@bar[2]{\if@single{#1}{\wide@bar@{#1}{#2}{1}}{\wide@bar@{#1}{#2}{2}}}
\newcommand*\wide@bar@[3]{%
  \begingroup
  \def\mathaccent##1##2{%
    \let\mathaccent\save@mathaccent
    \if#32 \let\macc@nucleus\first@char \fi
    \setbox\z@\hbox{$\macc@style{\macc@nucleus}_{}$}%
    \setbox\tw@\hbox{$\macc@style{\macc@nucleus}{}_{}$}%
    \dimen@\wd\tw@
    \advance\dimen@-\wd\z@
    \divide\dimen@ 3
    \@tempdima\wd\tw@
    \advance\@tempdima-\scriptspace
    \divide\@tempdima 10
    \advance\dimen@-\@tempdima
    \ifdim\dimen@>\z@ \dimen@0pt\fi
    \rel@kern{0.6}\kern-\dimen@
    \if#31
      \overline{\rel@kern{-0.6}\kern\dimen@\macc@nucleus\rel@kern{0.4}\kern\dimen@}%
      \advance\dimen@0.4\dimexpr\macc@kerna
      \let\final@kern#2%
      \ifdim\dimen@<\z@ \let\final@kern1\fi
      \if\final@kern1 \kern-\dimen@\fi
    \else
      \overline{\rel@kern{-0.6}\kern\dimen@#1}%
    \fi
  }%
  \macc@depth\@ne
  \let\math@bgroup\@empty \let\math@egroup\macc@set@skewchar
  \mathsurround\z@ \frozen@everymath{\mathgroup\macc@group\relax}%
  \macc@set@skewchar\relax
  \let\mathaccentV\macc@nested@a
  \if#31
    \macc@nested@a\relax111{#1}%
  \else
    \def\gobble@till@marker##1\endmarker{}%
    \futurelet\first@char\gobble@till@marker#1\endmarker
    \ifcat\noexpand\first@char A\else
      \def\first@char{}%
    \fi
    \macc@nested@a\relax111{\first@char}%
  \fi
  \endgroup
}
\makeatother

\newcommand{\ouralgtext}{{GRPO\texttt{-}DT}\xspace}
\newcommand{\ourppotext}{{PPO\texttt{-}DT}\xspace}

\newcommand{\grpomath}{\mathsf{GRPO}\xspace}
\newcommand{\ppomath}{\mathsf{PPO}\xspace}
\newcommand{\odttdthree}{{ODT\texttt{+}TD3}\xspace}
\newcommand{\odt}{\text{ODT}\xspace}

\newcommand{\rtg}{\text{RTG}\xspace}

\newcommand{\rtgmath}{\mathsf{RTG}\xspace}
\newcommand{\tdthree}{\text{TD3}\xspace}
\newcommand{\replay}{\mathsf{replay}\xspace}
\newcommand{\traj}{\mathsf{traj}\xspace}

\newcommand{\old}{\mathsf{old}\xspace}
\newcommand{\refmath}{\mathsf{ref}\xspace}
\newcommand{\sub}{\mathsf{sub}\xspace}
\newcommand{\eval}{\mathsf{eval}\xspace}
\newcommand{\initialmath}{\mathsf{initial}\xspace}
\newcommand{\train}{\mathsf{train}\xspace}
\newcommand{\policymath}{\mathsf{policy}\xspace}
\newcommand{\valuemath}{\mathsf{value}\xspace}
\newcommand{\lrmath}{\mathsf{lr}\xspace}
\newcommand{\batchmath}{\mathsf{batch}\xspace}
\newcommand{\onlinemath}{\mathsf{online}\xspace}
\newcommand{\offlinemath}{\mathsf{offline}\xspace}
\newcommand{\actualmath}{\mathsf{actual}\xspace}
\newcommand{\cltrain}{\mathsf{CL_{train}}\xspace}

\title{Online Finetuning Decision Transformers with Pure RL Gradients}
\date{}
\author{
Junkai Luo\\
{\normalsize University of California, Riverside}\\
{\normalsize\texttt{junkail@ucr.edu}}
\and
Yinglun Zhu\textsuperscript{\dag}\\
{\normalsize University of California, Riverside}\\
{\normalsize\texttt{yzhu@ucr.edu}}
}

\begin{document}

\maketitle

\begingroup
\renewcommand\thefootnote{}
\footnotetext{\textsuperscript{\dag}Project lead and corresponding author.}
\endgroup

\begin{abstract}

Decision Transformers (DTs) have emerged as a powerful framework for sequential decision making by formulating offline reinforcement learning (RL) as a sequence modeling problem. However, extending DTs to online settings with \emph{pure RL gradients} remains largely unexplored, as existing approaches continue to rely heavily on supervised sequence-modeling objectives during online finetuning. We identify hindsight return relabeling---a standard component in online DTs---as a critical obstacle to RL-based finetuning: while beneficial for supervised learning, it is fundamentally incompatible with importance sampling-based RL algorithms such as GRPO, leading to unstable training. Building on this insight, we propose new algorithms that enable online finetuning of Decision Transformers using pure reinforcement learning gradients. We adapt GRPO to DTs and introduce several key modifications, including sub-trajectory optimization for improved credit assignment, sequence-level likelihood objectives for enhanced stability and efficiency, and active sampling to encourage exploration in uncertain regions. Through extensive experiments, we demonstrate that our methods outperform existing online DT baselines and achieve new state-of-the-art performance across multiple benchmarks, highlighting the effectiveness of pure-RL-based online finetuning for Decision Transformers.

\end{abstract}
\section{Introduction}

The transformer architecture \citep{vaswani2017attention} lies at the core of modern foundation models.
Large language models (LLMs), in particular, have demonstrated remarkable generalization and reasoning capabilities through a simple yet powerful paradigm: large-scale pretraining followed by supervised and reinforcement learning (RL)-based finetuning \citep{radford2018improving,brown2020language,ouyang2022training,achiam2023gpt,comanici2025gemini}.
Inspired by this success, the \emph{Decision Transformer} (DT, \citealp{chen2021decision}) introduces transformers to sequential decision making by reframing classical RL as a conditional sequence modeling problem.
As an \emph{offline} RL method, DTs are trained with a supervised objective on pre-collected trajectories, effectively performing imitation learning \citep{hussein2017imitation} while conditioning on return-to-go (RTG) tokens.
\looseness=-1

The \emph{Online Decision Transformer} (\odt, \citealp{zheng2022online}) extends DTs to the online setting by enabling finetuning after offline pretraining.
During online finetuning, ODT collects new trajectories and applies \emph{hindsight return relabeling}, replacing intended RTGs with realized returns to align trajectories with their outcomes, mirroring the offline training procedure.
Subsequent work further augments ODT with TD3 \citep{fujimoto2018addressing} gradients, resulting in \odttdthree \citep{yan2024reinforcement}.
Despite these advances, existing approaches to online finetuning of DTs remain dominated by supervised objectives: \odt relies exclusively on supervised loss, while \odttdthree assigns only a small weight to RL gradients.

In contrast, recent progress in LLM finetuning shows that \emph{pure RL} methods—such as Group Relative Policy Optimization (GRPO, \citealp{shao2024deepseekmath})—can substantially improve reasoning and alignment \citep{guo2025deepseek,team2025qwq}.
This contrast naturally raises an important question:

\begin{center}
\textit{Can Decision Transformers be finetuned online using pure RL gradients?}
\end{center}

To investigate this question, we revisit the training paradigm of existing online DT methods and identify a fundamental limitation.
We show that \emph{hindsight return relabeling} \citep{zheng2022online}, while beneficial for supervised learning, is fundamentally incompatible with on-policy RL algorithms that rely on importance sampling.
Relabeling RTGs used during rollout with returns observed afterward introduces a mismatch in importance ratios, leading to unstable optimization and degraded performance (\cref{fig:hindsightrelabel}).
Removing hindsight return relabeling is therefore a necessary first step toward applying importance sampling-based algorithms, such as GRPO and PPO, to online finetuning of Decision Transformers.

Building on this insight, we develop a new framework for online finetuning of DTs using \emph{pure RL gradients}.
Specifically, we adapt GRPO to Decision Transformers (\ouralgtext) and introduce three key modifications:
(i) a sub-trajectory-based optimization objective that enables fine-grained credit assignment, supported by either environment resetting \citep{mhammedi2024power,kazemnejad2025vineppo} or an auxiliary Q-function;
(ii) a sequence-level importance ratio that improves training stability and efficiency; and
(iii) an active sampling strategy that prioritizes uncertain states to enhance exploration.
Together, these components enable RL-only finetuning of pretrained DTs and yield new state-of-the-art performance across multiple benchmarks.
In addition, we adapt Proximal Policy Optimization (PPO, \citealp{schulman2017proximalpolicyoptimizationalgorithms}) to Decision Transformers (\ourppotext), achieving competitive results where prior PPO-based approaches failed \citep{yan2024reinforcement}.

\paragraph{Contributions.}
Our main contributions are summarized as follows:
\begin{enumerate}[label=(\roman*)]
\item We identify {hindsight return relabeling}—a standard component in existing online DT methods—as a critical obstacle to reinforcement learning-based finetuning: while effective for supervised objectives, it is incompatible with importance sampling-based policy gradient algorithms such as GRPO and PPO.
\item We propose \ouralgtext, an adaptation of GRPO for Decision Transformers that enables \emph{pure-RL online finetuning} by integrating sub-trajectory optimization for improved credit assignment, sequence-level importance ratios for enhanced stability and efficiency, and active state sampling for better exploration. These modifications are general and may be of independent interest beyond DTs.
\item 
We conduct extensive experiments and demonstrate that pure RL finetuning---via \ouralgtext and \ourppotext---outperforms existing online DT baselines and achieves new state-of-the-art performance across multiple benchmarks.

\end{enumerate}

\paragraph{Paper organization.}
The remainder of the paper is organized as follows.
\cref{sec2:preliminaries} reviews background and preliminaries.
\cref{sec3:method} presents our methods.
\cref{sec4:experiment} reports experimental results.
\cref{sec5:relatedwork} discusses related work, and \cref{sec6:conclusion} concludes the paper.

\section{Preliminaries}
\label{sec2:preliminaries}

  \paragraph{Markov Decision Process.} We formulate the reinforcement learning environment as a \textit{Markov Decision Process} (MDP), defined by a tuple $\cM = (\mathcal{S}, \mathcal{A}, P, R)$. Here, $\mathcal{S}$ is the state space, $\mathcal{A}$ is the action space, $P(s_{h+1} \mid s_h, a_h)$ is the probability transition function, $R(s_h, a_h)$ is the reward function. 
  At each timestep $h = 1, \dots, H$, the agent observes a state $s_h \in \mathcal{S}$, selects an action $a_h \in \mathcal{A}$ according to a policy $\pi(a_h \mid s_h)$, transitions to the next state $s_{h+1} \sim P(\cdot \mid s_h, a_h)$, and receives a reward $r_h = R(s_h, a_h)$. 
  The goal is to learn a policy $\pi$ that maximizes the expected cumulative reward
$\mathbb{E}_{\mathcal{M}, \pi}\!\left[\sum_{h=1}^{H} r_h\right]$.

  \paragraph{Decision Transformer (DT).} Decision Transformer \citep{chen2021decision} represents a powerful paradigm for \emph{offline} reinforcement learning, formulating decision making as a sequence modeling problem with pre-collected training trajectories. 
  A DT trajectory consists of three types of tokens: {return-to-go} (RTG), state, and action, where the \rtg $g_h \ldef \sum_{\wb h = h}^{H} r_{\wb h}$ represents the sum of future reward from step $h$ onward.
  DT leverages the transformer architecture \citep{vaswani2017attention} to autoregressively learn a policy from pre-collected trajectories. {DTs are trained on fixed-length trajectory segments \citep{chen2021decision}:} let $m$ denotes the context length, the DT learns to generate the next action $a_h$ based on past interactions $( (g, s,a)_{h-m+1: h-1}, g_h, s_h ) \ldef (g_{h-m+1}, s_{h-m+1}, a_{h-m+1}, \dots, g_h, s_h)$ of context length $m$.
  The model is trained via supervised learning by minimizing the mean squared error (MSE) between the predicted action $\pi(a_h \mid ( (g, s,a)_{h-m+1: h-1}, g_h, s_h ))$ and the ground-truth action $a_h$.
  When the context is clear, we use the shorthand $\pi(a_h \mid s_h, g_h) \ldef \pi(a_h \mid ( (g, s,a)_{h-m+1: h-1}, g_h, s_h ))$. 
  During evaluation and deployment, the learner specifies a desired initial RTG, since the ground-truth future RTG isn't known in advance, and leverages the DT to autoregressively generate the next action and interact with the environment.

  \paragraph{Online finetuning of Decision Transformers.} Online Decision Transformer (\odt, \citealp{zheng2022online}) extends offline DT to the \emph{online} setting by first conducting offline pretraining and then online finetuning with interactively collected data. 
  The offline pretraining stage largely follows the standard DT training procedure.
  In the online finetuning stage, the DT is deployed into the environment with a desired initial RTG $g_{\onlinemath}$ to collect new trajectories that gradually replacing old trajectories stored in the replay buffer; the replay buffer is initialized with the offline trajectories.
  For online collected trajectories, \odt applies \emph{hindsight return relabeling} \citep{andrychowicz2017hindsight, ghosh2019learning} to relabel the RTG tokens based on the actual achieved RTG $g_{\actualmath}$.
  \odt adopts a stochastic Gaussian policy to account for exploration in the online setting.
  However, similar to offline DT, \odt still learns via a supervised learning objective of minimizing the negative log-likelihood loss.

  While \odt improves model performance during online finetuning, recently, \citet{yan2024reinforcement} pointed out that \odt fails in settings with medium or low-quality offline data due to the sole use of the supervised learning objective. The supervised learning objective learns  $\frac{\partial a}{\partial \rtgmath}$, i.e., how action changes as the target RTG varies, since DT models actions conditioned on RTGs.
  However, what actually drives online policy improvement is $\frac{\partial \rtgmath}{\partial a}$, i.e., how RTG responds to action adjustments, especially when offline pretraining data is not of high-quality; see section 3.1 in \citet{yan2024reinforcement} for more details.
  To enable better online improvements, \citet{yan2024reinforcement} propose \odttdthree, which augments the supervised \odt objective (with hindsight return relabeling) with RL gradients from \tdthree \citep{fujimoto2018addressing, fujimoto2021minimalist} to guide online exploration and adaptation.
  However, \odttdthree still prioritizes the supervised \odt loss, assigning only a small weight to the RL gradients during optimization.

  \paragraph{Group Relative Policy Optimization (GRPO).} 
  GRPO is a reinforcement learning algorithm initially proposed for large language models (LLMs) finetuning \citep{shao2024deepseekmath, guo2025deepseek}.
  It simplifies Proximal Policy Optimization (PPO, \citealp{schulman2017proximalpolicyoptimizationalgorithms}) by removing the need for a value model to estimate the advantages.
  Instead, GRPO samples multiple responses per question and uses their average reward as the baseline for advantage calculation.
  Specifically, for each query $q \sim \Delta(Q)$ sampled from the question distribution $\Delta(Q)$, the model generates a group of $G$ responses $\crl{o_1, \cdots, o_G}$ based on policy $\pi_{\theta_\old}$.
  A reward $r_i$ is computed for each response $o_i$, usually with the help of a reward model.
  GRPO optimizes the policy model by maximizing the following objective:
  \begin{align}   
    J_{\grpomath}(\theta) &= 
    \mathbb{E}_{q \sim {\Delta(Q)}, 
    \{o_i\}_{i=1}^G \sim \pi_{\theta_\old (\cdot \mid q)}
  } \nonumber \\ 
    & \quad \frac{1}{G} \sum_{i=1}^G 
    \frac{1}{|o_i|} \sum_{h=1}^{|o_i|} 
    \min 
    \left( 
        w_{i,h}(\theta) \wh {A}_i, \, 
        \mathsf{clip} \left( 
            w_{i,h}(\theta), 
            1 - \epsilon, 
            1 + \epsilon 
        \right) \wh {A}_i \right) 
        - \beta D_{\mathsf{KL}} \left( \pi_\theta \parallel \pi_{\theta_\refmath} \right),
  \label{eq:originegrpoadv}
  \end{align}
  where  
$\wh {A}_i = \frac{r_i - \mathrm{mean}(\{r_1, r_2, \cdots, r_G\})}{\mathrm{std}(\{r_1, r_2, \cdots, r_G\})}$ denotes the advantage of the $i$-th rollout, 
  $w_{i,h}(\theta)= \frac{\pi_\theta (o_{i,h} | q, o_{i,<h})}{\pi_{\theta_{\old}} (o_{i,h} | q, o_{i,<h})}$ denotes the importance sampling ratio, and $D_{\mathsf{KL}} ( \pi_\theta \parallel \pi_{\theta_\refmath} )$ denotes the KL penalty that prevents large deviations from a reference policy $\pi_{\theta_\refmath}$.
  In our implementation,
  we employ a slow update schedule for the reference policy $\pi_{\theta_{\refmath}}$, updating it once every four updates of $\pi_\theta$.
  \looseness=-1

\section{Methods}
\label{sec3:method}

In \cref{sec:3.1}, we first analyze a key limitation of adapting Decision Transformers to the online setting using importance-sampling-based reinforcement learning algorithms, and discuss how this limitation can be addressed. Building on these insights, \cref{sec:3.2} presents our adaptation of GRPO to Decision Transformers, incorporating several key modifications. We introduce additional extensions in \cref{sec:extensions}.

  \subsection{Removing Hindsight Return Relabeling}
  \label{sec:3.1}
  When deploying DTs, the learner must specify a desired initial RTG, since the ground-truth future RTG is unknown in advance.  
In Online Decision Transformer (\odt), the learner typically sets a relatively high target RTG $g_{\onlinemath}$ during rollout to encourage optimistic exploration.  
During training, a key component of \odt---known as \emph{hindsight return relabeling}---replaces the RTG tokens based on the actual achieved RTG $g_{\actualmath}$ \citep{zheng2022online}.  

When augmenting the supervised \odt objective with RL gradients from \tdthree, \odttdthree \citep{yan2024reinforcement} also adopt the hindsight return relabeling step.  
\citet{yan2024reinforcement} further attempt to perform online finetuning of DTs using PPO \citep{schulman2017proximalpolicyoptimizationalgorithms}---an importance sampling-based RL algorithms---but find that PPO gradients lead to poor performance, ultimately reverting to \tdthree\ gradients instead.

\begin{figure}[t]
  \centering
    \includegraphics[width=0.24\textwidth]{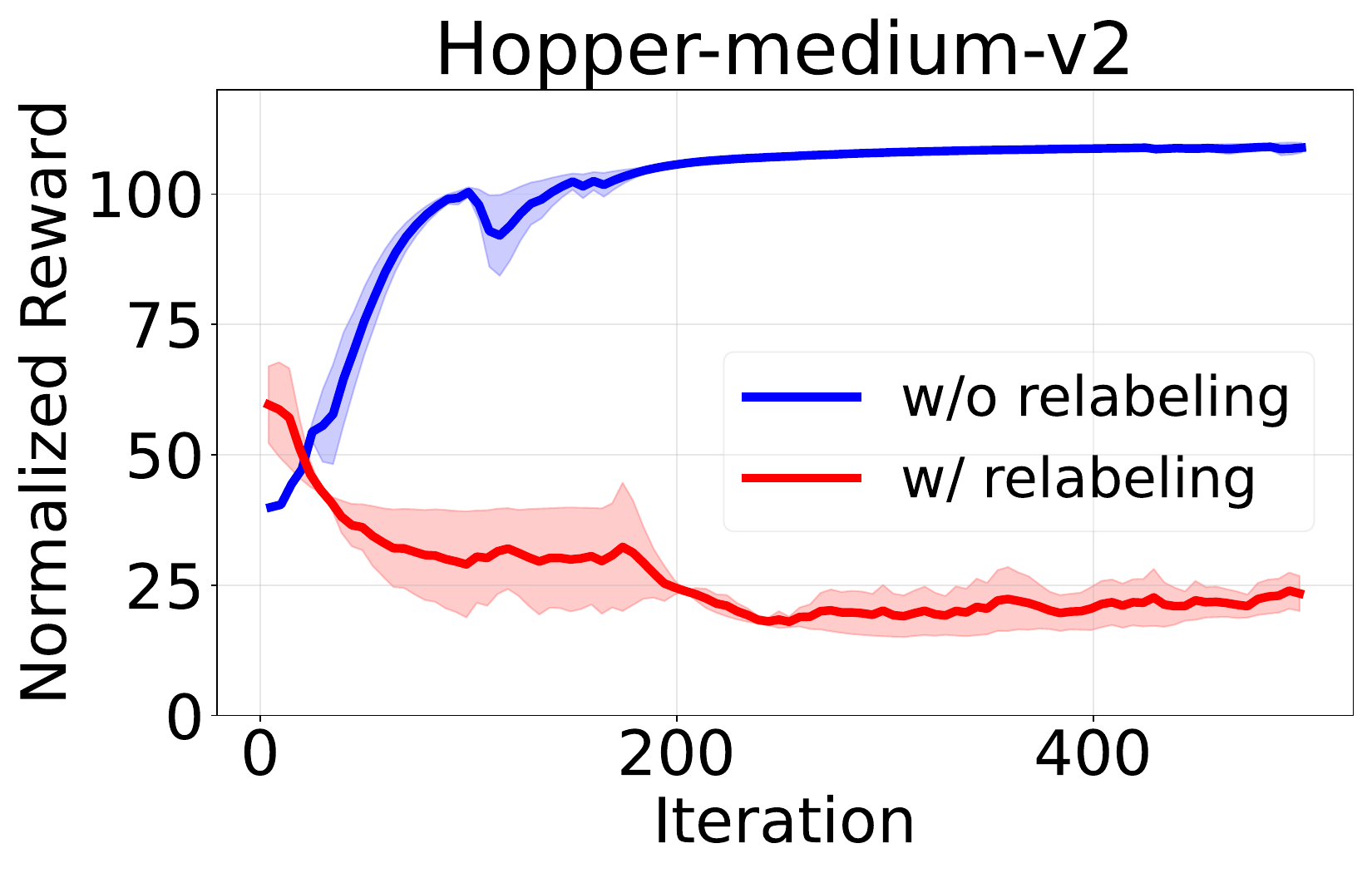}
    \includegraphics[width=0.24\textwidth]{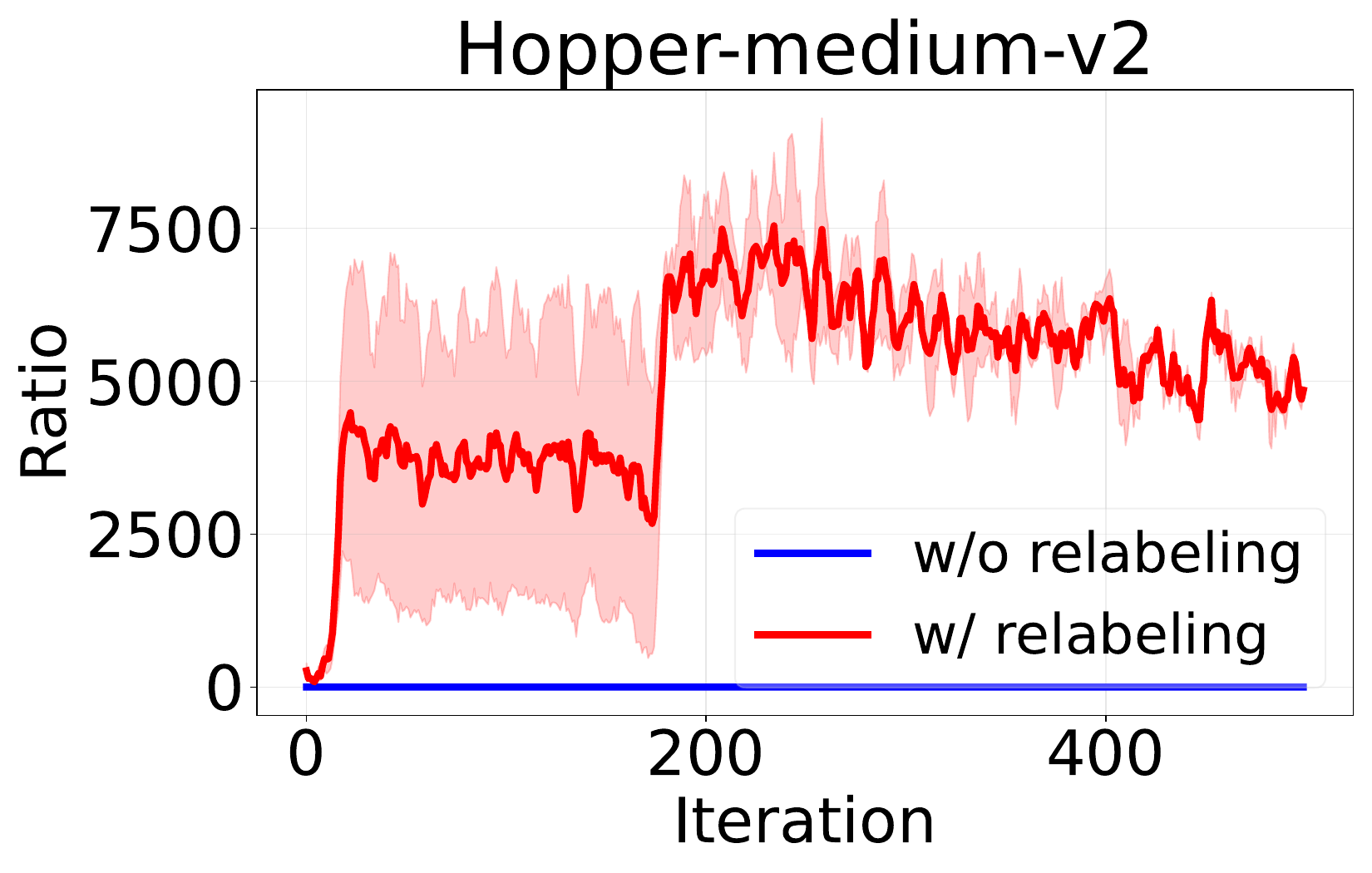}
    \includegraphics[width=0.24\textwidth]{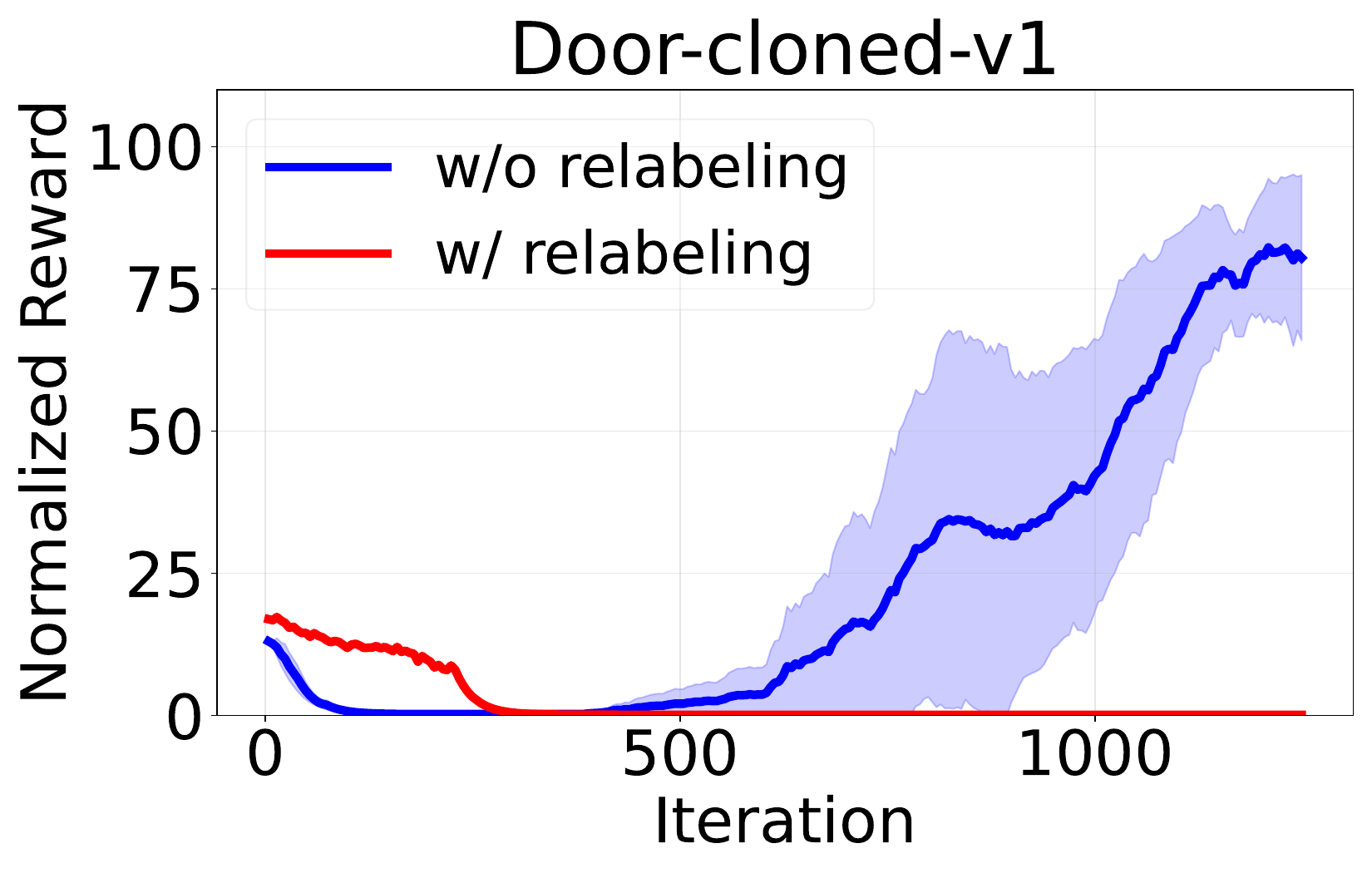}
    \includegraphics[width=0.24\textwidth]{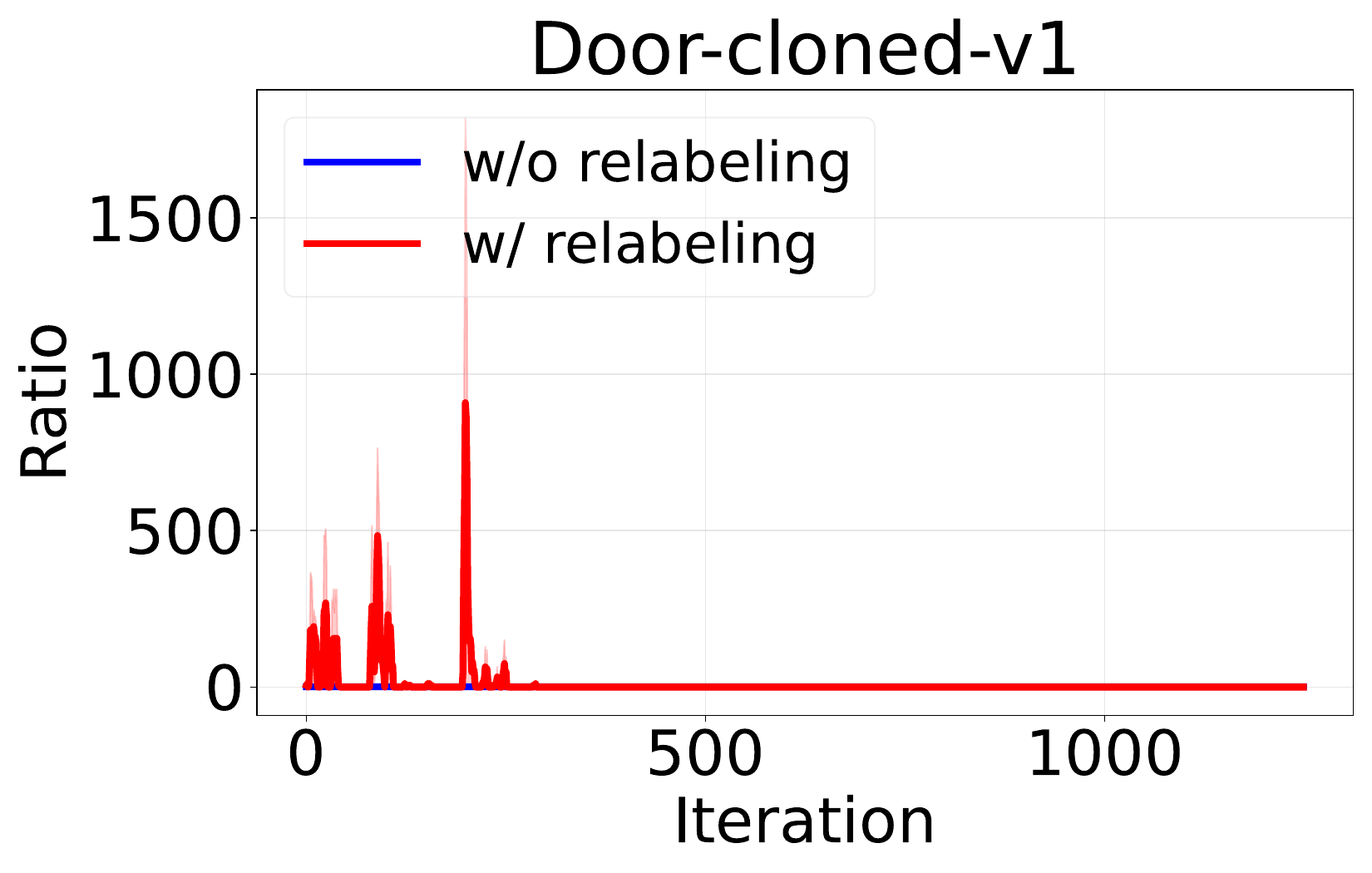}
  \caption{
Comparison of GRPO-DT with and without hindsight return relabeling when applied to sampled sub-trajectories.
Hindsight return relabeling causes the importance-sampling ratio
  $\frac{\pi_\theta(a \mid s, g)}{\pi_{\theta_{\text{old}}}(a \mid s, g)}$
  to become highly unstable, which in turn degrades performance.}
  \label{fig:hindsightrelabel}
\end{figure}

While hindsight return relabeling works well under supervised learning objectives (see Fig.~5.4 in \citet{zheng2022online}), we find it incompatible with importance sampling-based RL gradients that rely on the ratio $\frac{\pi_\theta(a \mid s, g)}{\pi_{\theta_\old}(a \mid s, g)}$, as in PPO and GRPO.  
The issue arises from a mismatch in the conditioning variable $g$: the learner conditions rollouts on a high RTG $g_{\onlinemath}$ for optimistic exploration, yet the achieved RTG $g_{\actualmath}$ is often much lower.  
If hindsight return relabeling is applied, actions sampled from $\pi_{\theta_\old}(a \mid s, g_{\onlinemath})$ are later trained as if they were drawn from $\pi_{\theta_\old}(a \mid s, g_{\actualmath})$, producing unreliable importance weights and unstable updates.  
This inconsistency explains why naive applications of PPO to \odt\ tend to fail \citep{yan2024reinforcement}. 
As shown in \cref{fig:hindsightrelabel}, removing hindsight return relabeling significantly improves stability and overall performance.  
In simpler environments such as MuJoCo Hopper, relabeling may yield transient gains but eventually leads to collapse, whereas in more complex environments such as Adroit Door, the model fails to learn altogether when relabeling is enabled.

Additionally, hindsight return relabeling requires access to the returns of the entire trajectory, which introduces additional challenges when optimizing over sub-trajectory rollouts (introduced in \cref{sec:3.2}). 
In the experiments shown in \cref{fig:hindsightrelabel}, we apply hindsight return relabeling to sub-trajectories by rolling out the full trajectory.
\looseness=-1

\subsection{Adapting GRPO to Decision Transformers} 
\label{sec:3.2}

\begin{algorithm}[t]
  \caption{Online Finetuning Decision Transformers with GRPO (\ouralgtext)}

\label{alg:grpowdt}
	\renewcommand{\algorithmicrequire}{\textbf{Input:}}	\renewcommand{\algorithmicensure}{\textbf{Output:}}
\begin{algorithmic}[1]
\REQUIRE
  Pretrained policy $\pi_{\theta_1}$, full trajectory buffer $\mathcal{T}_\replay$, sub-trajectory buffer $\mathcal{T}_\mathsf{sub}$, number of iterations $T$, initial \rtg $g_\onlinemath$, number of reset points in a trajectory $K$, sub-trajectory length $L_\traj$, evaluation steps $L_\eval$, group size $G$ for GRPO.

\FOR{iteration $t = 1, \cdots, T$}
\label{line:iteration}
    \STATE Roll out a full trajectory $\tau$ using the current policy $\pi_{\theta_t}(\cdot \mid s_1, g_\onlinemath)$, conditioned on (randomized) initial state $s_1$ and \rtg $g_\onlinemath$; 
    update $\mathcal{T}_{\text{replay}}$ with $\tau$. \algcommentlight{Collect full trajectory; FIFO buffer update.}
    \vspace{-10 pt}
    \STATE Sample a minibatch $\mathcal{B}$ of full trajectories from $\mathcal{T}_\replay$ from distribution $p$ with $p(\tau) \ldef \frac{|\tau|}{\sum_{\tau \in \mathcal{T}} |\tau|}$.
    \FOR {each full trajectory $\tau \in \mathcal{B}$}
    \STATE Sample $K$ reset points $\crl{(s_{k}, g_k)}_{k=1}^K$ from action-variance distribution.
    \STATE For each reset point $(s_{k}, g_k)$, generate $G$ sub-trajectories $\{\tau^{\sub}_{k_i}\}_{i=1}^G$ of length $L_{\traj}$ with the current policy $\pi_{\theta_t}$; evaluate each sub-trajectory for $L_\eval$ more steps to get reward $R(\tau^{\sub}_{k_i})$.
    \algcommentlight{Sub-trajectory generation and evaluation.}
    \label{line:reset}
        \STATE Compute the advantage $\wh A_{k_i}$ for each sub-trajectory $\tau_{k_i}^\sub$ using \cref{eq:normalization}. 
    \STATE Update the sub-trajectory buffer $\cT_{\sub}$ with $\crl{(\tau_{k_i}^\sub, \wh A_{k_i}, (s_k, g_k) )}_{i=1}^{|G|}$.
    \algcommentlight{FIFO buffer update.}
    \ENDFOR
    \STATE Finetune the current policy with sub-trajectories in $\cT_\sub$ using the sequence-level importance ratio (\cref{eq:grpo_advantage}) to get a new policy $\pi_{\theta_{t+1}}$.
\ENDFOR
  \ENSURE Online finetuned policy $\pi_{\theta_{T+1}}$.

\end{algorithmic}
\end{algorithm}

\begin{figure}[t]
  \centering
    \begin{minipage}{0.48\textwidth}
    \centering
    \includegraphics[width=0.48\textwidth]{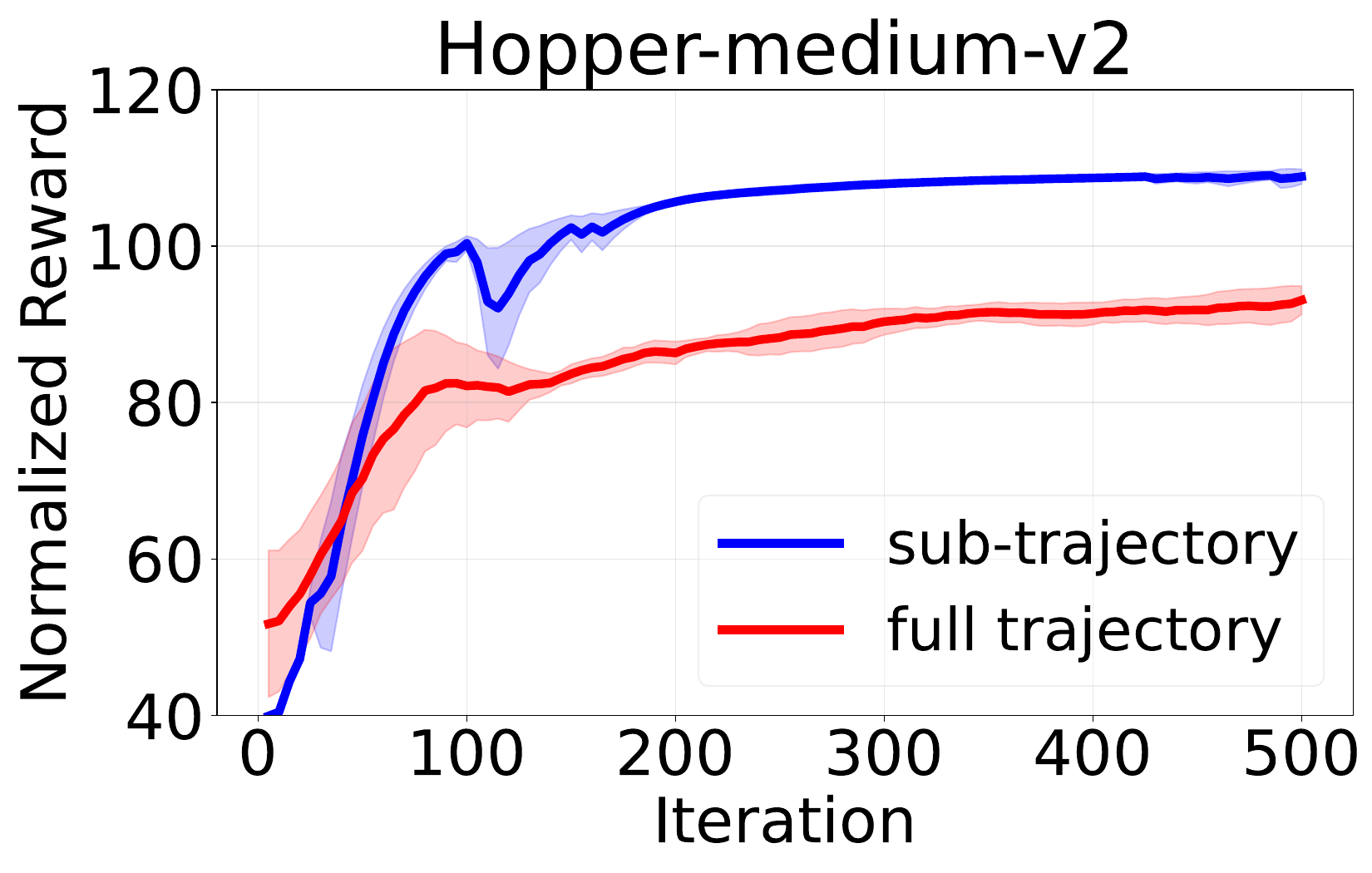}
    \includegraphics[width=0.48\textwidth]{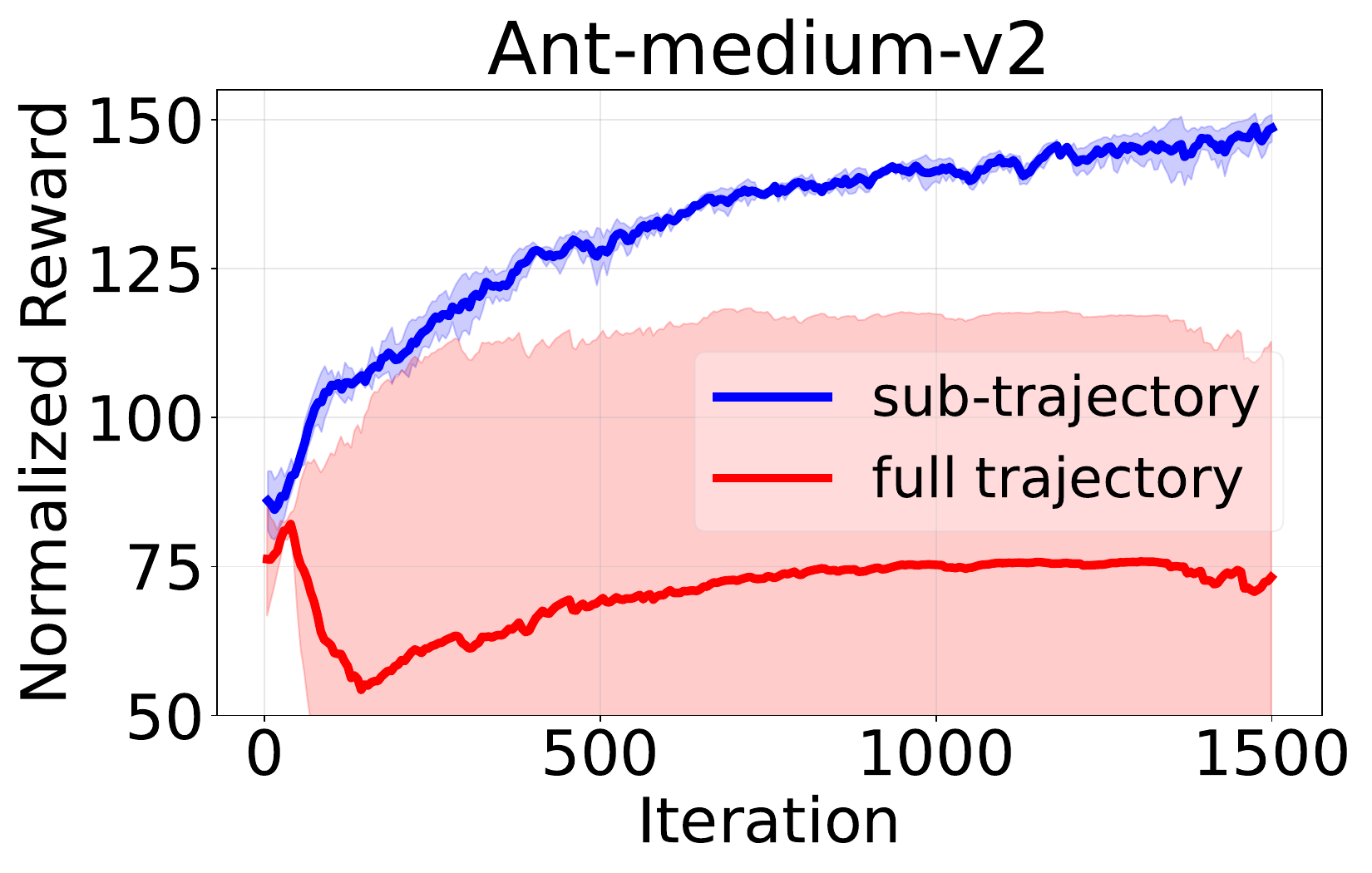}
    \subcaption{Sub-trajectory vs.\ full trajectory}
    \label{subfig:credit-assignment}
  \end{minipage}\hfill
  \begin{minipage}{0.48\textwidth}
    \centering
    \includegraphics[width=0.48\textwidth]{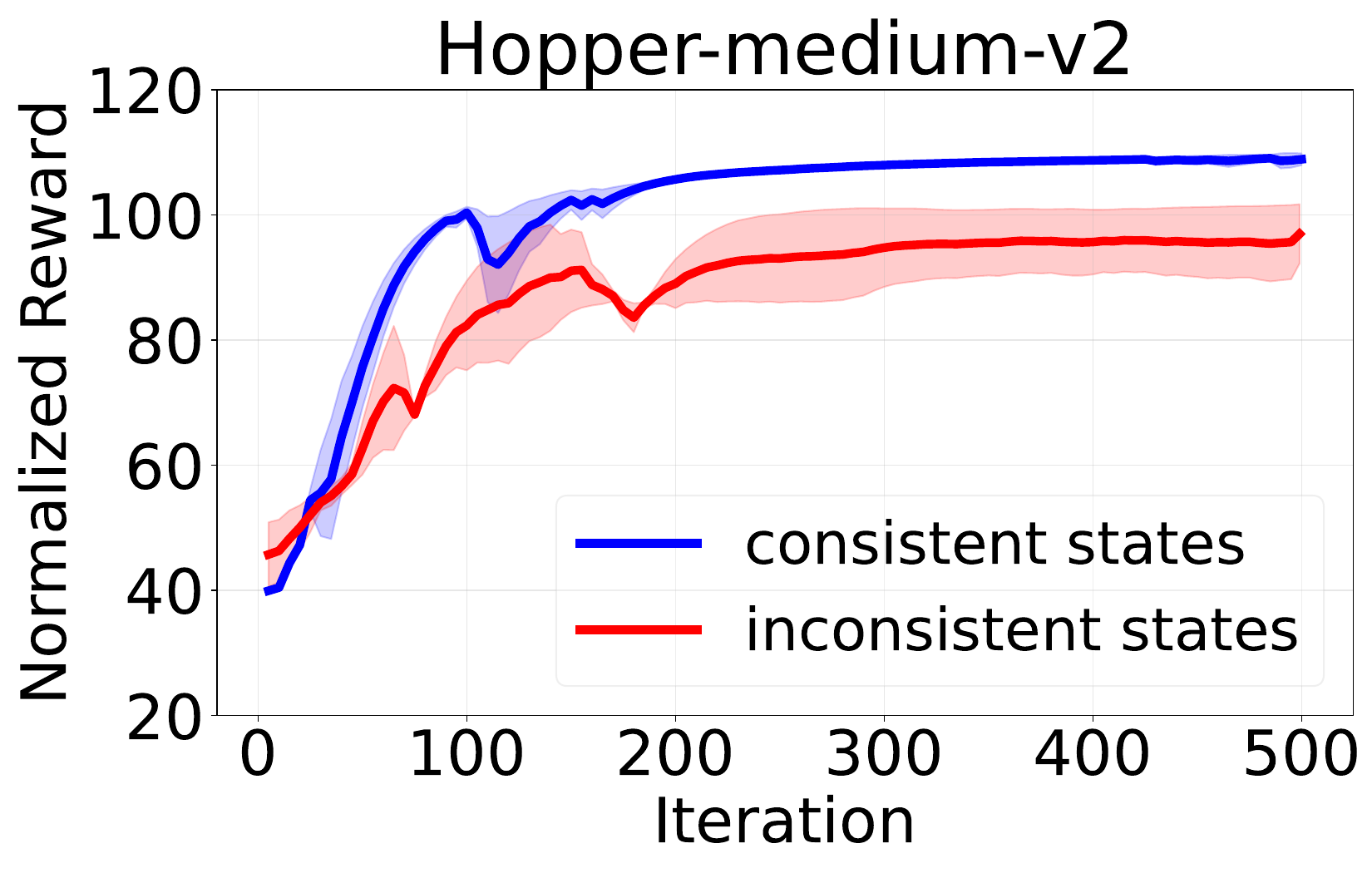}
    \includegraphics[width=0.48\textwidth]{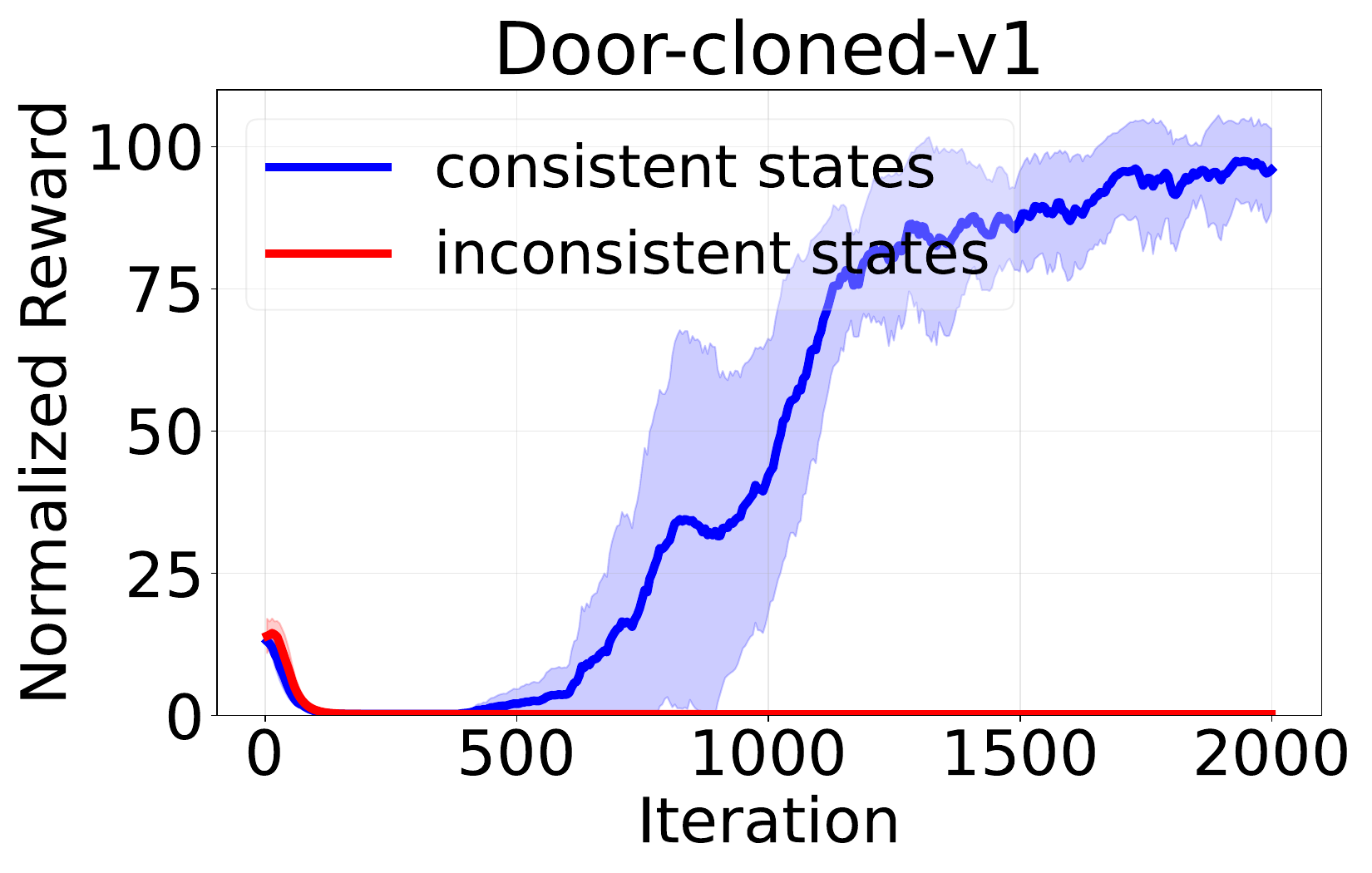}
    \subcaption{Consistent vs.\ inconsistent states}
    \label{subfig:consistent-states}
  \end{minipage}
  
  \begin{minipage}{0.48\textwidth}
    \centering
    \begin{minipage}{0.48\textwidth}
      \centering
      \includegraphics[width=\textwidth]{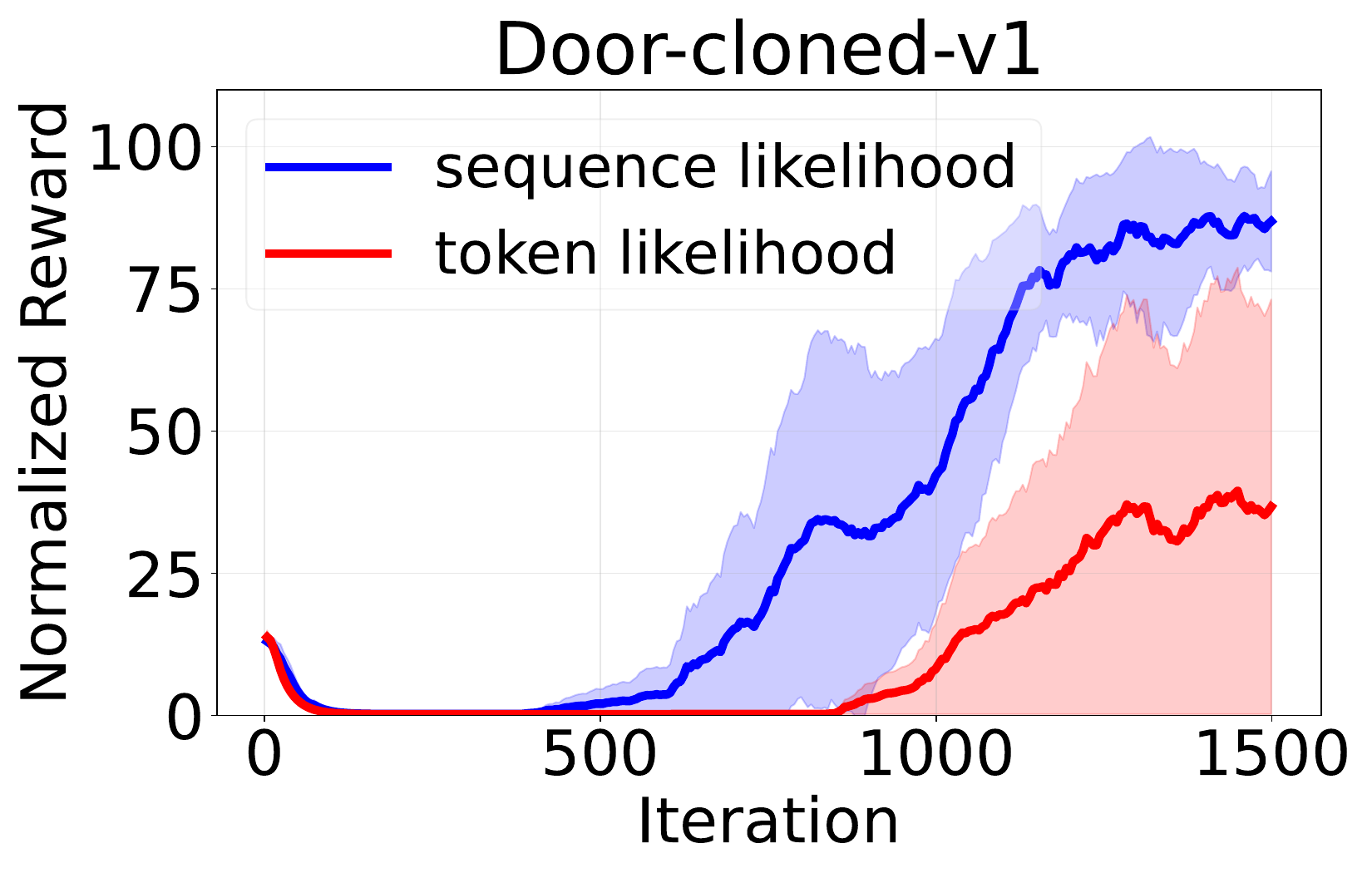}
      \subcaption{Sequence vs.\ token}
      \label{subfig:sequencelikelihood}
    \end{minipage}
    \begin{minipage}{0.48\textwidth}
      \centering
      \includegraphics[width=\textwidth]{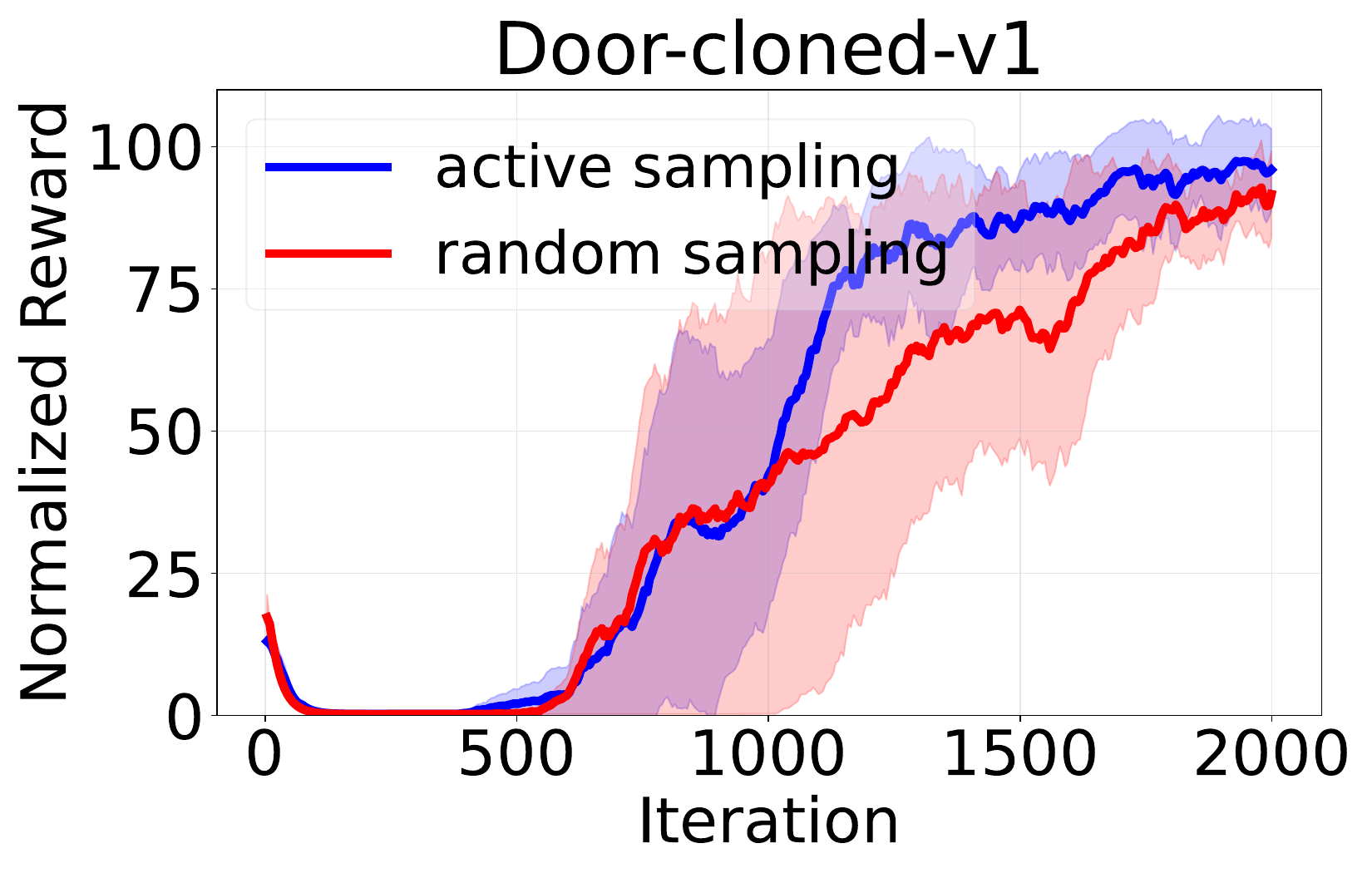}
      \subcaption{Active vs.\ random}
      \label{subfig:activeselection}
    \end{minipage}
  \end{minipage}\hfill
  \begin{minipage}{0.48\textwidth}
    \centering
    \includegraphics[width=0.48\textwidth]{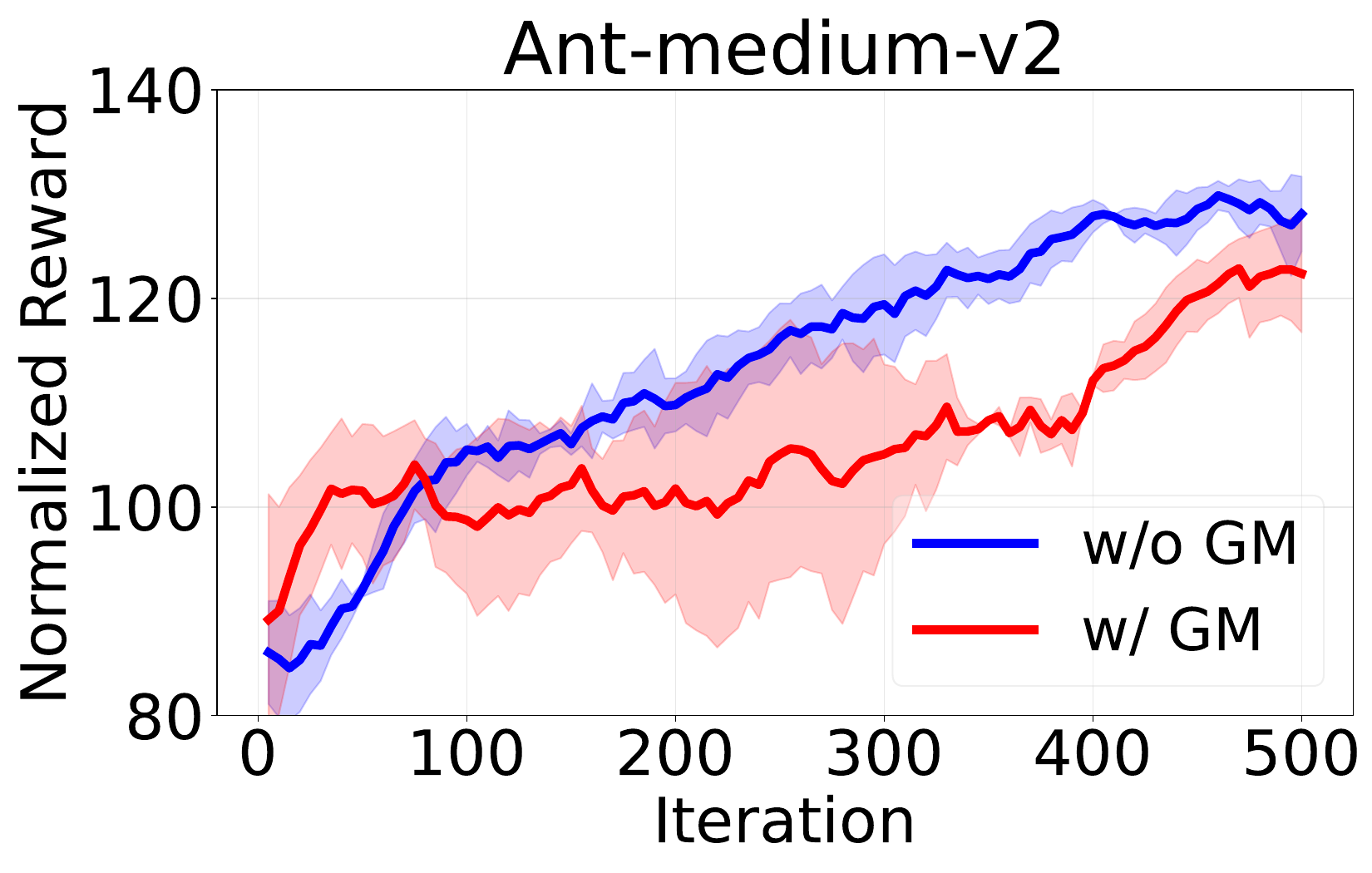}
    \includegraphics[width=0.48\textwidth]{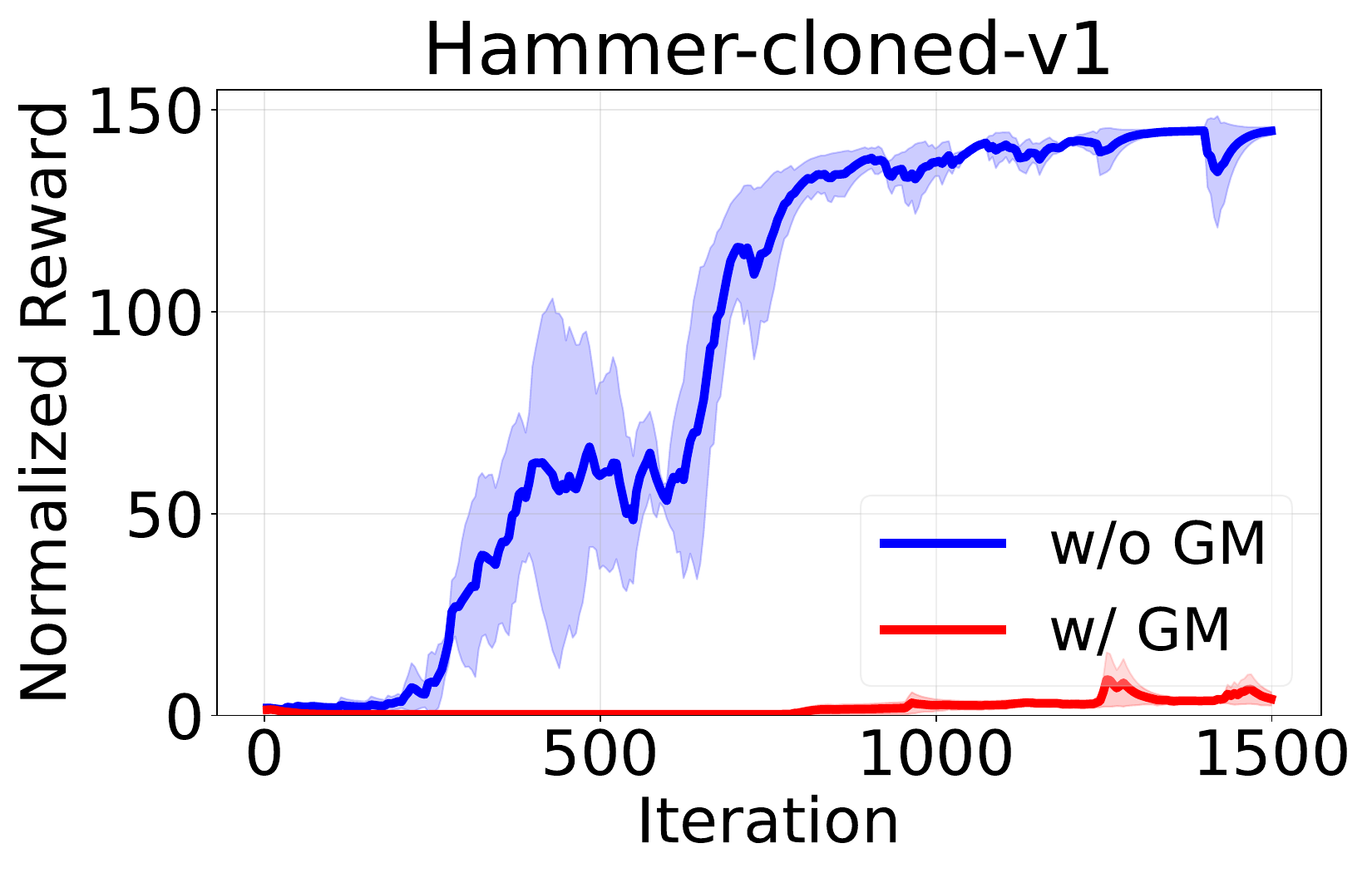}
    \subcaption{Removing geometric mean in sequence-level ratio}
    \label{subfig:geometricmean}
  \end{minipage}
  \caption{
  Ablation studies illustrating the impact of key design choices in GRPO-DT.
We compare variants of GRPO-DT with and without several proposed components across multiple environments.
(a) Optimizing over sub-trajectories leads to faster convergence and higher final performance compared to updating on full trajectories.
(b) Sampling groups from consistent states significantly improves learning stability over mixing inconsistent states.
(c) Computing importance ratios using sequence-level likelihoods yields more reliable updates than token-wise likelihoods.
(d) Actively selecting reset points based on uncertainty accelerates learning compared to random sampling.
(e) Removing geometric mean (GM) normalization improves performance when computing sequence-level importance ratios in the sub-trajectory regime.
  }
  \label{fig:ablation}
\end{figure}

We provide an overview of online finetuning Decision Transformers with GRPO in \cref{alg:grpowdt} (\ouralgtext),
which is achieved by optimization on \emph{sub-trajectories} rather than full trajectories as used in the original GRPO formulation \citep{shao2024deepseekmath, guo2025deepseek}.  
At each iteration, the current policy interacts with the environment to collect complete rollouts, from which we sample reset points and generate groups of sub-trajectories for each reset point.
The reset points are also \emph{actively selected} based on the variance in the policy action distribution.
The algorithm computes advantages for each sub-trajectory according to \cref{eq:normalization}.
These sub-trajectories and their advantages are then used to update the policy with a \emph{sequence-level importance ratio} described in \cref{eq:grpo_advantage}.

Compared to vanilla GRPO, our method introduces three key design modifications to better align GRPO with Decision Transformers.  
Specifically, (i) we redesign the optimization objective to operate on a group of sub-trajectories rather than full rollouts, enabled by either resetting or learning an extra Q-function;  
(ii) we compute importance weights at the \emph{sequence level} to better align computed advantages;  
and (iii) we incorporate an \emph{active sampling} mechanism that prioritizes uncertain states for optimization.  
We describe each of these design choices in detail below and also provide abalations to demonstrate their effectiveness.

\paragraph{Sub-trajectory rollouts for better credit assignment.}
In its original formulation for language models, GRPO assigns a single response-level reward to all tokens within the same sequence \citep{shao2024deepseekmath, guo2025deepseek}, thereby discarding fine-grained credit assignment.  
A straightforward adaptation to classical RL environments and tasks would aggregate all stepwise rewards in a rollout and assign this trajectory-level return uniformly to every timestep.  
However, such a formulation performs poorly in RL environments: as shown in \cref{subfig:credit-assignment}, the model fails to learn when trained with full trajectories in the Ant-medium-v2 environment.  
This limitation is expected, as RL tasks---particularly those in continuous control---require more precise credit assignment than language modeling.
Whereas tokens in a sentence tend to be coherently correlated, actions in RL can lead to drastically different outcomes (e.g., distinct action choices when navigating a maze).

To address this limitation, we adapt GRPO for Decision Transformers using a \emph{sub-trajectory formulation}.  
We first select $K$ reset points $\crl{(s_k, g_k)}_{k=1}^K$ from each full trajectory.  
For every reset point $(s_{k}, g_k)$, we generate $G$ sub-trajectories $\crl{\tau^\sub_{k_i}}_{i=1}^{|G|}$ of length $L_\traj$ using the current policy $\pi_{\theta_t}$. To better attribute rewards to each sub-trajectory, we further roll out each one for an additional $L_\eval$ steps using the expected action (or the most probable action in the discrete case).
The reward $r^\sub_{k_i}$ for sub-trajectory $\tau^\sub_{k_i}$ is defined as the cumulative discounted reward over $L_\traj + L_\eval$ steps, with a discount factor $\gamma$ emphasizing rewards obtained near the reset point. 
For the additional $L_{\eval}$ steps, we use the expected action to reduce the variance induced by stochastic action sampling, resulting in more stable sub-trajectory return estimates.
We then compute the advantage for each sub-trajectory in its group as:
\begin{align}
  \wh A_{k_i} =
  \frac{r_{k_i}^\sub - \mathrm{mean}(\{r_{k_1}^\sub, r_{k_2}^\sub, \dots, r_{k_{|G|}}^\sub\})}
       {\mathrm{std}(\{r_{k_1}^\sub, r_{k_2}^\sub, \dots, r_{k_{|G|}}^\sub\})}.
  \label{eq:normalization}
\end{align}
Only the sub-trajectory of length $L_\traj$ is used for GRPO optimization, while the subsequent $L_\eval$ steps are used solely for evaluation.
The parameter $L_\traj$ controls the granularity of credit assignment, whereas $L_\eval$ determines the quality of reward estimation.
Empirically, we find that a relatively small $L_\traj$ combined with a relatively large $L_\eval$ yields the best performance; see \cref{sec:ablation} for detailed ablations on these hyperparameters.

To ensure stable optimization, we enforce \emph{state consistency} by resetting vectorized environments to the same states before generating sub-trajectories within each group.  
This reset mechanism is essential for convergence, as shown in \cref{subfig:consistent-states}.  
In scenarios where environment resetting is infeasible, we train an auxiliary Q-function using \tdthree~\citep{fujimoto2018addressing} to evaluate candidate actions under a shared state (see \cref{sec:qgrpo} for details).  
This Q-function-guided variant also achieves competitive performance, as demonstrated in \cref{sec:ablation}.

\paragraph{Sequence-level importance ratio.}
In standard GRPO, importance weights are computed at the token level, reflecting per-step likelihoods.  
When adapting GRPO to Decision Transformers, we find that computing importance weights at the \emph{sequence level}---that is, over entire sub-trajectories of length $L_\traj$---significantly improves model performance.  
Intuitively, the sequence-level importance ratio 
$\frac{\pi_\theta(\tau^\sub_{k_i} \mid s_k, g_k)}{\pi_{\theta_{\old}}(\tau^\sub_{k_i} \mid s_k, g_k)}$
is better aligned with the advantage $\wh A_{k_i}$, which is already computed at the sequence level.
Empirical validation in \cref{subfig:sequencelikelihood} further supports this intuition.

We modify the original GRPO objective to 
incorporate sequence-level importance weighting.
Since the sub-trajectory buffer $\cT_\sub$ stores sub-trajectory $\tau^\sub_{k_i}$ along with its advantage $\wh A_{k_i}$ and reset point $(s_k, g_k)$, we randomly sample sub-trajectories from the sub-trajectory buffer and maximize the following objective: 
\begin{align}
    J_{\grpomath}(\theta) &= 
    \mathbb{E}_{({\tau^{\sub}_{k_i}}, \wh A_{k_i}, (s_k, g_k)) \sim \unif(\cT_\sub)}
   \nonumber \\  
   & \quad
    \min\Bigg(
        \frac{\pi_\theta(\tau^\sub_{k_i} \mid s_k, g_k)}
             {\pi_{\theta_{\old}}(\tau^\sub_{k_i} \mid s_k, g_k)} \wh A_{k_i}, \,
        \mathsf{clip}\!\left(
            \frac{\pi_\theta(\tau^\sub_{k_i} \mid s_k, g_k)}
                 {\pi_{\theta_{\old}}(\tau^\sub_{k_i} \mid s_k, g_k)}, 
            1 - \epsilon, 
            1 + \epsilon 
        \right) \wh A_{k_i}
    \!\Bigg)
    - \beta D_{\mathsf{KL}}\!\left( \pi_\theta \, \| \, \pi_{\theta_\refmath} \right).
    \label{eq:grpo_advantage}
\end{align}
This sequence-level objective yields more stable and sample-efficient optimization, consistent with concurrent findings by \citet{zheng2025group} in language modeling.  
While \citet{zheng2025group} propose using a geometric mean over sequence-level importance ratios, we find that removing geometric normalization (as in \cref{eq:grpo_advantage}) leads to superior performance when combined with sub-trajectory optimization using relatively short rollout lengths (\cref{subfig:geometricmean}). 
We hypothesize that this improvement arises because learning without geometric normalization (i) more rapidly suppresses outdated sub-trajectories through clipping, and (ii) permits more aggressive updates for approximately on-policy sub-trajectories.

While maximizing the GRPO objective in \cref{eq:grpo_advantage}, we impose an entropy constraint
$H(\pi_\theta(\tau_{k_i}^\sub \mid s_k, g_k)) \ge \rho$
to encourage exploration. Here,
$H(\pi_\theta(\tau_{k_i}^\sub \mid s_k, g_k))$
denotes the average policy entropy computed over the sub-trajectory $\tau_{k_i}^\sub$, and $\rho$ is a predefined lower bound.
Following \citet{zheng2022online}, we relax this hard constraint using a primal-dual formulation
$\min_{\kappa \ge 0}\; \max_{\theta}\;
J(\theta) + \kappa\big(H(\pi_\theta(\tau_{k_i}^\sub \mid s_k, g_k)) - \rho\big)$,
where $\kappa$ is the dual variable.
In practice, this corresponds to augmenting \cref{eq:grpo_advantage} with an additional entropy term
$\kappa H(\pi_\theta(\tau_{k_i}^\sub \mid s_k, g_k))$,
while adaptively updating $\kappa$ during training to ensure that the entropy constraint
$H(\pi_\theta(\tau_{k_i}^\sub \mid s_k, g_k)) \ge \rho$
is approximately satisfied.

  \paragraph{Active sampling for state selection.}
During policy rollouts, we observe that certain state-RTG pairs $(s_h, g_h)$ exhibit higher variance in the predicted action distribution $\pi_\theta(\cdot \mid s_h, g_h)$. 
Since higher variance indicates greater predictive uncertainty, resetting the starting points of sub-trajectory rollouts to such uncertain states can promote more effective exploration and accelerate learning. 
Motivated by this observation, we introduce an \emph{active sampling} mechanism that biases sub-trajectory selection toward high-uncertainty states.

Specifically, for each trajectory $\tau$, we compute a scalar uncertainty score at each step $h$ by aggregating the diagonal covariance of the action distribution.
Given $\pi_\theta(\cdot\mid s_h,g_h)=\mathcal N(\mu_\theta,\Sigma_\theta)$ with $\Sigma_\theta=\mathrm{diag}(\sigma_{h,1}^2,\dots,\sigma_{h,d}^2)$.
We define
$\sigma_h^2 \;:=\; \frac{1}{d}\sum_{i=1}^d \sigma_{h,i}^2$,
and apply a softmax transformation to obtain the sampling distribution
$p_h := \frac{\exp(\sigma_h^2)}{\sum_{h'=1}^{|\tau|}\exp(\sigma_{h'}^2)}$.
Sub-trajectory reset points are then sampled according to $p$.
This active selection strategy prioritizes learning updates in regions of high uncertainty, leading to faster and more stable convergence (\cref{subfig:activeselection}).
\looseness=-1

\subsection{Extensions}
\label{sec:extensions}

\subsubsection{Q-Guided \ouralgtext}
\label{sec:qgrpo}

In this section, we introduce \emph{Q-guided \ouralgtext}, a variant of \cref{alg:grpowdt} designed for settings in which environment resetting is infeasible. 
Q-guided \ouralgtext largely follows the pseudocode in \cref{alg:grpowdt}, except for the reset operation on line~\ref{line:reset}. 
Instead of explicitly rolling out sub-trajectories from a reset point $(s_h, g_h)$, we train an auxiliary Q-function $Q_{\phi_t}$ to evaluate actions sampled from the current policy $\pi_{\theta_t}(\cdot \mid s_h, g_h)$ for advantage computation.

Specifically, at each state-RTG pair $(s_h, g_h)$, we randomly sample a group of $G$ candidate actions
$\{a_{h_i}\}_{i=1}^G \sim \pi_{\theta_t}(\cdot \mid s_h, g_h)$.
Each sampled action is treated as defining a \emph{hypothetical sub-trajectory} that starts from $(s_h, g_h)$ by first executing $a_{h_i}$ and then following the policy $\pi_{\theta_t}$.
Under this interpretation, we evaluate the quality of each action using the Q-function and obtain a scalar reward
$R(a_{h_i}) = Q_{\phi_t}(s_h, a_{h_i})$.
These rewards are subsequently used to compute the normalized advantage following \cref{eq:normalization}.
The Q-function $Q_{\phi_t}$ is trained following the TD3 algorithm~\citep{fujimoto2018addressing}.

\subsubsection{Adapting PPO to Decision Transformers}
In addition to adapting GRPO to DTs, we further extend \textit{Proximal Policy Optimization} (PPO, \citealp{schulman2017proximalpolicyoptimizationalgorithms}) to the Decision Transformer framework, denoted as \ourppotext.
Unlike prior attempts that directly apply PPO to DTs \citep{yan2024reinforcement}, our approach removes the \emph{hindsight return relabeling} step during online finetuning, as discussed in \cref{sec:3.1}.
This adjustment ensures consistency between the rollout and training objectives, which is crucial for stable optimization under importance sampling.
Our formulation also differs from PPO adaptations in multi-agent reinforcement learning that omit return-to-go (RTG) conditioning entirely \citep{meng2023offline}, effectively reducing DTs to behavior cloning \citep{ross2010efficient,hussein2017imitation}.
Following \citet{zheng2022online}, we perform optimization on sub-trajectories sampled from full trajectories and incorporate an adaptive entropy term to encourage exploration.
Since PPO computes advantages at the token level, \ourppotext\ accordingly employs \textit{token-level importance ratios} instead of sequence-level ratios used in our \ouralgtext adaptation.
For training the value network, we leverage generalized advantage estimation (GAE, \citealp{schulman2015high}) that combine Monte Carlo estimates with temporal-difference bootstrapping, yielding smoother and more stable value function updates.
Further implementation details of \ourppotext\ are provided in \cref{sec:ppo}.

\section{Experiments}
\label{sec4:experiment}

In this section, we empirically evaluate our online finetuning algorithms for Decision Transformers using pure RL gradients.
We describe the experimental setup in \cref{sec:exp_setup}, present the main results in \cref{sec:exp_main}, and provide additional analyses and ablations in \cref{sec:ablation}.

\subsection{Experimental Setups}
\label{sec:exp_setup}

\textbf{Environments and datasets.} 
We evaluate methods on three continuous control and manipulation environments from D4RL \citep{fu2020d4rl}:  
(i) \textbf{MuJoCo} \citep{todorov2012mujoco} tasks, including \textit{Hopper}, \textit{Walker2d}, and \textit{Ant}, with dense rewards, evaluated on the \textit{medium}, \textit{medium-replay}, and \textit{random} datasets.  
(ii) \textbf{Adroit} manipulation tasks \citep{rajeswaran2017learning}, including \textit{Door}, \textit{Hammer}, and \textit{Pen}, evaluated on the \textit{human} and \textit{cloned} datasets.  
(iii) \textbf{AntMaze} \citep{fu2020d4rl} with sparse goal-reaching rewards (a reward of 1 if success and 0 otherwise), using the \textit{umaze} and \textit{umaze-diverse} datasets.  
Detailed descriptions of each environment and dataset are provided in \cref{sec:envs&datasets}.

\textbf{Baselines.}
We compare our adapted \ouralgtext and \ourppotext against three main baselines:
(i) \textbf{Online Decision Transformer} (\textbf{ODT}; \citet{chen2021decision}), the standard online extension of DT that uses a supervised loss as the finetuning objective;
(ii) \textbf{\odttdthree} \citep{yan2024reinforcement}, the current state-of-the-art method for online finetuning of Decision Transformers; and
(iii) \textbf{IQL} \citep{kostrikov2021offline}, a widely used offline RL algorithm for which we employ its \emph{online variant} in our experiments.
For reference, we also report the performance of the offline pretrained \textbf{Decision Transformer} (\textbf{DT}) without online finetuning.
Detailed hyperparameter settings are provided in \cref{sec:hyperparams}.

\textbf{Metrics.} 
Following D4RL~\citep{fu2020d4rl}, we report the normalized final reward for each algorithm, where higher values indicate better performance.
All results are averaged over three runs, with standard deviations reported.
We additionally present learning curves that track normalized reward throughout training.

For learning curves, the x-axis corresponds to the number of outer learning iterations, i.e., line~\ref{line:iteration} in \cref{alg:grpowdt} and line~3 of Algorithm~1 in \citet{zheng2022online}.
This choice reflects a compromise for unifying different classes of methods.
Specifically, IQL, ODT, and \odttdthree require nearly two orders of magnitude more gradient updates (and thus more computation) than our methods, \ouralgtext and \ourppotext, while our adapted policy-gradient-based algorithms consume several to tens of times more environment interactions.
This difference arises because IQL, ODT, and \odttdthree benefit from extensive experience replay, whereas policy gradient methods typically require more online interactions.

To facilitate a fair comparison of asymptotic performance, we train all algorithms for a larger number of iterations than those reported in \citet{yan2024reinforcement}.
Evaluation is conducted after the gradient updates of each iteration.
As a result, even at iteration~0, all methods have already undergone several updates, during which their behaviors may diverge and yield different outcomes.

\begin{table}[t]
\caption{
Comparison of average normalized final return of each method across various environments and datasets.  
The best results are shown in \textbf{bold}, and results within 10\% of the best are \underline{underlined}.  
Environment and task abbreviations are as follows:  
Ho = Hopper, Wa = Walker2d, An = Ant, D = Door, P = Pen, H = Hammer, U = UMaze, UD = UMaze-Diverse;  
dataset types: R = Random, M = Medium, MR = Medium-Replay, C = Cloned, H = Human.  
Each entry is reported as ``performance {\scriptsize $\pm$} standard deviation''. 
\looseness=-1
}

\label{tab:results}
\centering
\fontsize{7.5}{10}\selectfont
\begin{tabularx}{\textwidth}{l*{7}{>{\centering\arraybackslash}X}}
\toprule
  Environments & Datasets &DT &IQL & ODT & \odttdthree & \ourppotext & \ouralgtext \\
\midrule
\multirow{3}{1cm}{\shortstack{MuJoCo\\ (random)}}
&Ho-R-v2 &2.13 &41.02\,{\fontsize{6pt}{7pt}\selectfont $\pm$\,13.35} & 30.43\,{\fontsize{6pt}{7pt}\selectfont $\pm$\,0.01} & 83.32\,{\fontsize{6pt}{7pt}\selectfont $\pm$\,8.46} &  \textbf{106.97\,{\fontsize{6pt}{7pt}\selectfont $\pm$\,0.96}}  & \underline{99.20\,{\fontsize{6pt}{7pt}\selectfont $\pm$\,3.80}} \\
&Wa-R-v2 &4.53 &22.75\,{\fontsize{6pt}{7pt}\selectfont $\pm$\,1.56} & 10.88\,{\fontsize{6pt}{7pt}\selectfont $\pm$\,0.34}&82.95\,{\fontsize{6pt}{7pt}\selectfont $\pm$\,18.28} & \textbf{108.69\,{\fontsize{6pt}{7pt}\selectfont $\pm$\,8.86}} & \underline{100.25\,{\fontsize{6pt}{7pt}\selectfont $\pm$\,33.19}} \\
&An-R-v2  &31.41 &58.69\,{\fontsize{6pt}{7pt}\selectfont $\pm$\,23.03} & 19.08\,{\fontsize{6pt}{7pt}\selectfont $\pm$\,3.97} & 80.58\,{\fontsize{6pt}{7pt}\selectfont $\pm$\,7.25} & 107.45\,{\fontsize{6pt}{7pt}\selectfont $\pm$\,22.83} & \textbf{120.69\,{\fontsize{6pt}{7pt}\selectfont $\pm$\,5.47}} \\
\midrule
&Average &12.69 &40.82 &20.13& 82.28& \textbf{107.70} &\textbf{106.71}\\
\midrule
\multirow{6}{1cm}{\shortstack{MuJoCo\\(medium)}}
&Ho-M-v2 &46.46 & 74.19\,{\fontsize{6pt}{7pt}\selectfont $\pm$\,20.25} & \underline{98.02\,{\fontsize{6pt}{7pt}\selectfont $\pm$\,0.63}} & \underline{101.47\,{\fontsize{6pt}{7pt}\selectfont $\pm$\,2.29}} & \underline{105.65\,{\fontsize{6pt}{7pt}\selectfont $\pm$\,5.43}} & \textbf{108.81\,{\fontsize{6pt}{7pt}\selectfont $\pm$\,0.85}} \\
&\mbox{Ho-MR-v2} &38.12 &96.97\,{\fontsize{6pt}{7pt}\selectfont $\pm$\,2.16} & 87.73\,{\fontsize{6pt}{7pt}\selectfont $\pm$\,0.59} & \underline{107.94\,{\fontsize{6pt}{7pt}\selectfont $\pm$\,2.29}} & \textbf{109.60\,{\fontsize{6pt}{7pt}\selectfont $\pm$\,1.63}} & {83.61\,{\fontsize{6pt}{7pt}\selectfont $\pm$\,20.75}}\\ 
&Wa-M-v2 &47.95 &103.45\,{\fontsize{6pt}{7pt}\selectfont $\pm$\,1.37} & 76.49\,{\fontsize{6pt}{7pt}\selectfont $\pm$\,0.78} & 103.27\,{\fontsize{6pt}{7pt}\selectfont $\pm$\,5.95} &  109.49\,{\fontsize{6pt}{7pt}\selectfont $\pm$\,9.04}& \textbf{158.34\,{\fontsize{6pt}{7pt}\selectfont $\pm$\,3.75}}\\
&\mbox{Wa-MR-v2} &56.24 &103.00\,{\fontsize{6pt}{7pt}\selectfont $\pm$\,2.65} & 74.21\,{\fontsize{6pt}{7pt}\selectfont $\pm$\,2.41} & 102.80\,{\fontsize{6pt}{7pt}\selectfont $\pm$\,2.68} & 117.45\,{\fontsize{6pt}{7pt}\selectfont $\pm$\,14.79} & \textbf{137.36\,{\fontsize{6pt}{7pt}\selectfont $\pm$\,5.64}} \\
&An-M-v2 & 86.22 &118.18\,{\fontsize{6pt}{7pt}\selectfont $\pm$\,2.42} & 90.71\,{\fontsize{6pt}{7pt}\selectfont $\pm$\,0.03} & {131.56\,{\fontsize{6pt}{7pt}\selectfont $\pm$\,0.41}} & \underline{139.84\,{\fontsize{6pt}{7pt}\selectfont $\pm$\,0.95}} & \textbf{147.51\,{\fontsize{6pt}{7pt}\selectfont $\pm$\,2.44}}\\
&\mbox{An-MR-v2} &84.30  &117.51\,{\fontsize{6pt}{7pt}\selectfont $\pm$\,0.82} & 83.63\,{\fontsize{6pt}{7pt}\selectfont $\pm$\,0.87} & {120.01\,{\fontsize{6pt}{7pt}\selectfont $\pm$\,2.94}} & {117.95\,{\fontsize{6pt}{7pt}\selectfont $\pm$\,2.54}} & \textbf{142.05\,{\fontsize{6pt}{7pt}\selectfont $\pm$\,3.32}} \\
\midrule
&Average &59.88 &102.21 &85.13&111.18&\underline{116.66}&\textbf{129.61}\\
\midrule
\multirow{6}{1cm}{Adroit}
&D-C-v1 &0.14 & 46.72\,{\fontsize{6pt}{7pt}\selectfont $\pm$\,0.30}&1.26\,{\fontsize{6pt}{7pt}\selectfont $\pm$\,1.02} & 79.98\,{\fontsize{6pt}{7pt}\selectfont $\pm$\,5.62} & 0.19\,{\fontsize{6pt}{7pt}\selectfont $\pm$\,0.00} & \textbf{96.41\,{\fontsize{6pt}{7pt}\selectfont $\pm$\,7.59}}\\
&D-H-v1 &3.28 & 11.27\,{\fontsize{6pt}{7pt}\selectfont $\pm$\,0.44}&8.76\,{\fontsize{6pt}{7pt}\selectfont $\pm$\,3.87} & 79.73\,{\fontsize{6pt}{7pt}\selectfont $\pm$\,4.37} & \textbf{94.12\,{\fontsize{6pt}{7pt}\selectfont $\pm$\,3.99}} & \underline{89.33\,{\fontsize{6pt}{7pt}\selectfont $\pm$\,10.12}}\\
&P-C-v1 &43.31 &62.11\,{\fontsize{6pt}{7pt}\selectfont $\pm$\,13.25} & 16.24\,{\fontsize{6pt}{7pt}\selectfont $\pm$\,5.12} & \underline{109.86\,{\fontsize{6pt}{7pt}\selectfont $\pm$\,6.27}} &27.14\,{\fontsize{6pt}{7pt}\selectfont $\pm$\,0.24} &\textbf{111.15\,{\fontsize{6pt}{7pt}\selectfont $\pm$\,2.61}} \\
&P-H-v1 &33.10 &24.94\,{\fontsize{6pt}{7pt}\selectfont $\pm$\,1.48} & 19.84\,{\fontsize{6pt}{7pt}\selectfont $\pm$\,7.42} &\underline{77.18\,{\fontsize{6pt}{7pt}\selectfont $\pm$\,7.42}} &9.92\,{\fontsize{6pt}{7pt}\selectfont $\pm$\,5.00}  &\textbf{85.11\,{\fontsize{6pt}{7pt}\selectfont $\pm$\,6.08}}  \\
&H-C-v1 &0.65 & 4.87\,{\fontsize{6pt}{7pt}\selectfont $\pm$\,3.10} & 1.32\,{\fontsize{6pt}{7pt}\selectfont $\pm$\,0.06} & {119.95\,{\fontsize{6pt}{7pt}\selectfont $\pm$\,2.45}} & \underline{130.60\,{\fontsize{6pt}{7pt}\selectfont $\pm$\,2.81}} & \textbf{140.45\,{\fontsize{6pt}{7pt}\selectfont $\pm$\,1.93}}\\
&H-H-v1 &1.08 & 1.04\,{\fontsize{6pt}{7pt}\selectfont $\pm$\,1.56}&0.91\,{\fontsize{6pt}{7pt}\selectfont $\pm$\,0.22} & \underline{120.93\,{\fontsize{6pt}{7pt}\selectfont $\pm$\,2.18}} & \underline{129.23\,{\fontsize{6pt}{7pt}\selectfont $\pm$\,2.18}} & \textbf{132.64\,{\fontsize{6pt}{7pt}\selectfont $\pm$\,12.56}}\\
\midrule
&Average &13.59 &25.15 &8.06 & 97.93 & 65.2 & \textbf{109.18}\\ 
\midrule
\multirow{2}{1cm}{AntMaze}
&U-v2  &50.00 &91.21\,{\fontsize{6pt}{7pt}\selectfont $\pm$\,2.14} &89.27\,{\fontsize{6pt}{7pt}\selectfont $\pm$\,3.73}  &\textbf{99.64\,{\fontsize{6pt}{7pt}\selectfont $\pm$\,0.20}}  & 0.00\,{\fontsize{6pt}{7pt}\selectfont $\pm$\,0.00}&\underline{96.07\,{\fontsize{6pt}{7pt}\selectfont $\pm$\,0.53}} \\
&UD-v2  &65.00 &0.00\,{\fontsize{6pt}{7pt}\selectfont $\pm$\,0.00} &63.81\,{\fontsize{6pt}{7pt}\selectfont $\pm$\,1.64}  &\textbf{99.42\,{\fontsize{6pt}{7pt}\selectfont $\pm$\,0.43}}  &  47.00\,{\fontsize{6pt}{7pt}\selectfont $\pm$\,4.00}&\underline{97.70\,{\fontsize{6pt}{7pt}\selectfont $\pm$\,2.67}} \\
\midrule
&Average & 57.5 &45.61 & 76.54 & \textbf{99.53} &23.50 & \underline{96.89}\\
\bottomrule 
\end{tabularx}
\end{table}

\subsection{Main Results}
\label{sec:exp_main}

\cref{tab:results} reports the normalized returns and standard deviations averaged over three random seeds for each method.  
Overall, our proposed method \ouralgtext achieves the best performance across the majority of tasks, establishing new state of the arts.  
\ourppotext\ performs competitively in several cases but fails in certain environments (e.g., D-C-v1).  
\odttdthree\ achieves reasonable performance yet is generally outperformed by \ouralgtext.  
Both ODT and IQL underperform across most benchmarks, particularly on tasks with low-quality pretraining data such as the \textit{random} datasets, as well as on challenging environments like Adroit.  
It is worth noting that our implementation of \odttdthree\ uses longer training iterations (as described in \cref{sec:exp_setup}), leading to better results than those originally reported by \citet{yan2024reinforcement}.

\paragraph{Learning with random offline data.}
The first block of \cref{tab:results} evaluates model performance when pretrained on \textit{random} offline datasets \citep{fu2020d4rl} before the onset of online exploration.
The offline datasets consist of trajectories collected by an untrained policy and thus contain little meaningful supervision signal, serving as a standard stress test for offline-to-online adaptation and model robustness to poor pretraining quality \citep{yan2024reinforcement}.
\looseness=-1

Under this challenging setup, methods that rely primarily on supervised objectives struggle to improve during online finetuning: ODT fails to make meaningful progress.
IQL, while leveraging temporal-difference learning, suffers from inaccurate value estimation and limited policy improvement.
\odttdthree achieves moderate performance by augmenting supervised objectives with RL gradients, yet remains substantially weaker than our \ouralgtext\ and \ourppotext, which rely purely on reinforcement learning gradients.
By directly optimizing reward-driven RL signals, our methods effectively recover from poor initialization and achieve the best overall performance in this low-quality pretraining regime.

\paragraph{Learning with medium-quality offline data.}
The last three blocks of \cref{tab:results} report results obtained with medium-quality pretraining data before online exploration, representing realistic offline-to-online adaptation scenarios.
On the \textbf{MuJoCo} tasks, our adapted methods, \ouralgtext\ and \ourppotext, achieve the highest overall returns.
\odttdthree\ and IQL remain competitive, while \odt\ exhibits reasonable but inferior performance.

In the more challenging \textbf{Adroit} environment---where both state and action spaces are relatively high-dimensional---policies are prone to degradation or collapse during finetuning.
Under these conditions, ODT and IQL fail to improve beyond their pretrained performance, whereas our adapted \ouralgtext\ consistently achieves strong results, obtaining the best performance on 5 out of 6 datasets.
\odttdthree\ remains competitively on some Adroit tasks but is outperformed by our \ouralgtext across all datasets.
\ourppotext\ achieves strong results on certain tasks but shows limited improvement in others.

Finally, in the \textbf{AntMaze} environment with sparse, goal-reaching rewards (a reward of 1 for success and 0 otherwise), \odttdthree\ achieves the best performance, while our \ouralgtext\ follows closely---achieving returns within 4\% of the best score.
All other methods fail to make meaningful progress under this sparse-reward setting.
\looseness=-1

\begin{figure}[t]
  \centering
\includegraphics[width=\textwidth]{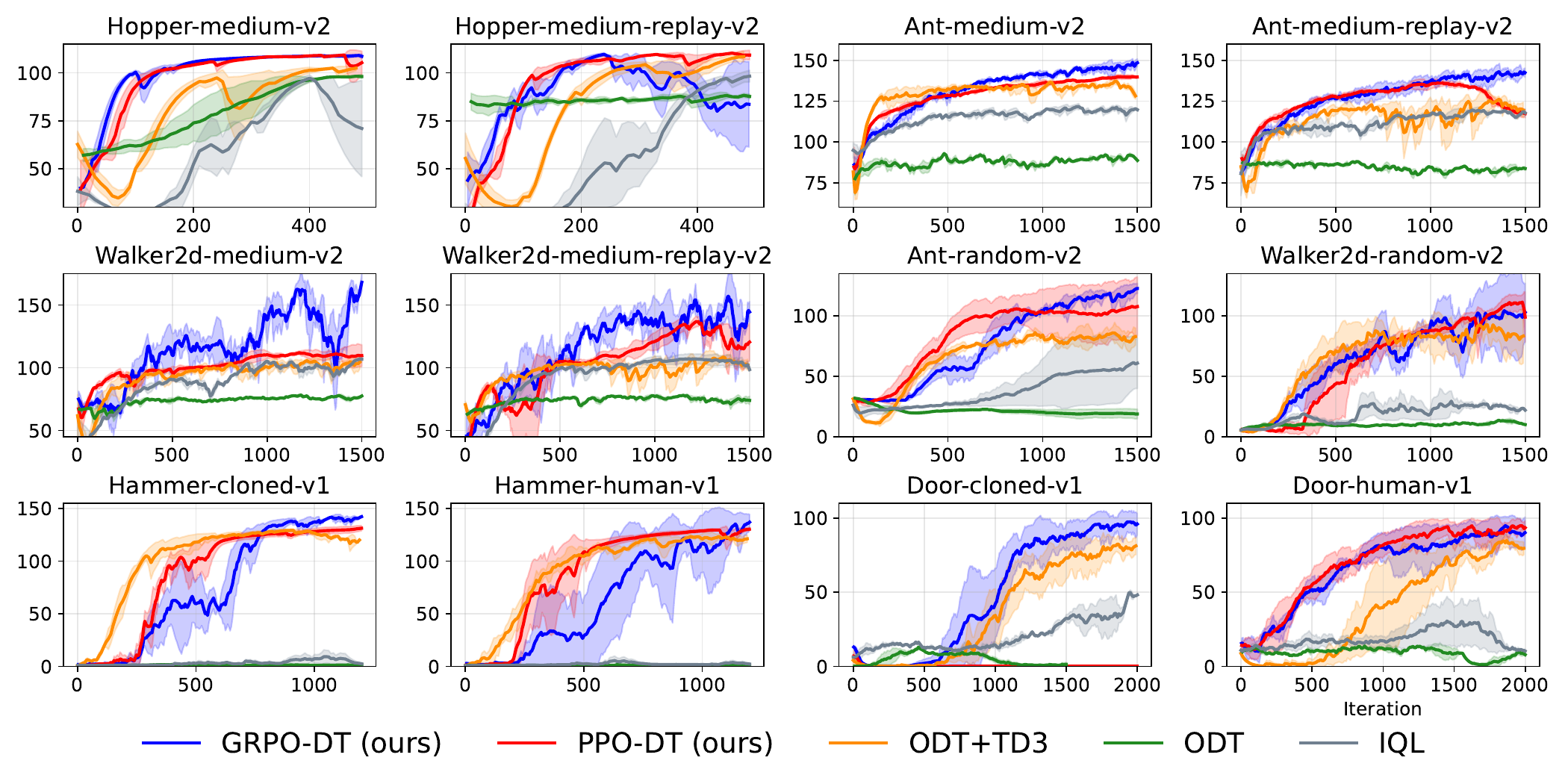}
\caption{Performance comparison across different RL environments.
Our proposed method \ouralgtext achieves the best performance across the majority of tasks.  
\ourppotext\ performs competitively in most cases but fails in certain environments.  
\odttdthree\ achieves overall decent performance but is generally outperformed by \ouralgtext.  
Both ODT and IQL consistently underperform across most environments.
}
\label{fig:exp_main}
\end{figure}

\paragraph{Computational and practical advantages over \odttdthree.}
Beyond achieving superior performance (as shown in \cref{tab:results,fig:exp_main}), our adapted \ouralgtext and \ourppotext also offer notable computational and practical advantages over \odttdthree.

\begin{enumerate}[label=(\roman*)]
\item \textbf{Fewer gradient updates.}
Sub-trajectory rollouts enable finer-grained credit assignment and more accurate gradient estimation, allowing our methods to learn effectively with far fewer updates.
For example, each iteration of our approach performs only $8 \times 256$ gradient updates, compared to roughly $256 \times 300$ updates required by \odttdthree\ (and \odt), representing a substantial reduction in compute cost.
\looseness=-1

\item \textbf{Seamless compatibility with pretrained DT-style models.}
Our method can directly finetune any pretrained DT-style model with minimal modification (see \cref{sec:reinformer} for experiments on other models), whereas \odttdthree\ requires \emph{modifying the offline pretraining objective} to integrate RL gradients and jointly train auxiliary Q-functions---significantly reducing its flexibility and scalability.

\item \textbf{Simpler implementation and improved stability.}
Unlike \odttdthree, which relies on auxiliary critic networks, our approach introduces no additional networks.
This critic-free design simplifies implementation, improves training stability, and facilitates reproducibility across different environments and pretrained models.

\end{enumerate}

\subsection{Additional Analyses and Ablations}
\label{sec:ablation}

We provide additional analyses and ablation studies in this section.
Empirical evidence supporting the key design choices in \cref{alg:grpowdt} is presented in \cref{fig:hindsightrelabel} and \cref{fig:ablation}, including the effects of removing hindsight return relabeling, using sub-trajectory rollouts, using consistent states for rollouts, adopting sequence-level importance ratios, applying active sampling for state selection, and removing geometric mean normalization for sequence-level importance ratios.

\begin{figure}[t]
  \centering
\begin{minipage}{0.24\textwidth}
    \centering
    \includegraphics[width=0.98\textwidth]{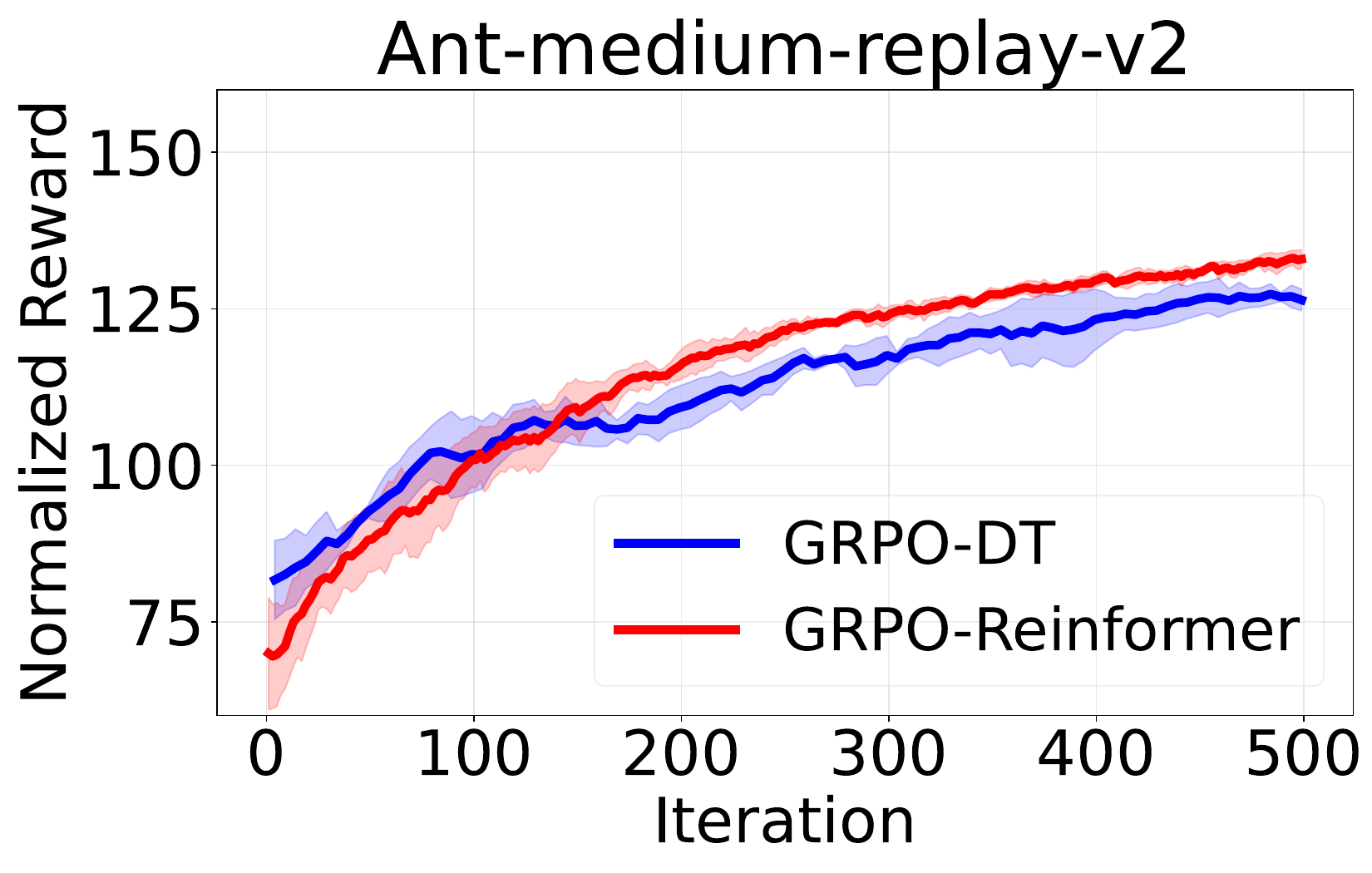}
\subcaption{GRPO on {Reinformer}}
\label{subfig:architectures}
\end{minipage}\hfill
\begin{minipage}{0.24\textwidth}
    \centering
    \includegraphics[width=0.98\textwidth]{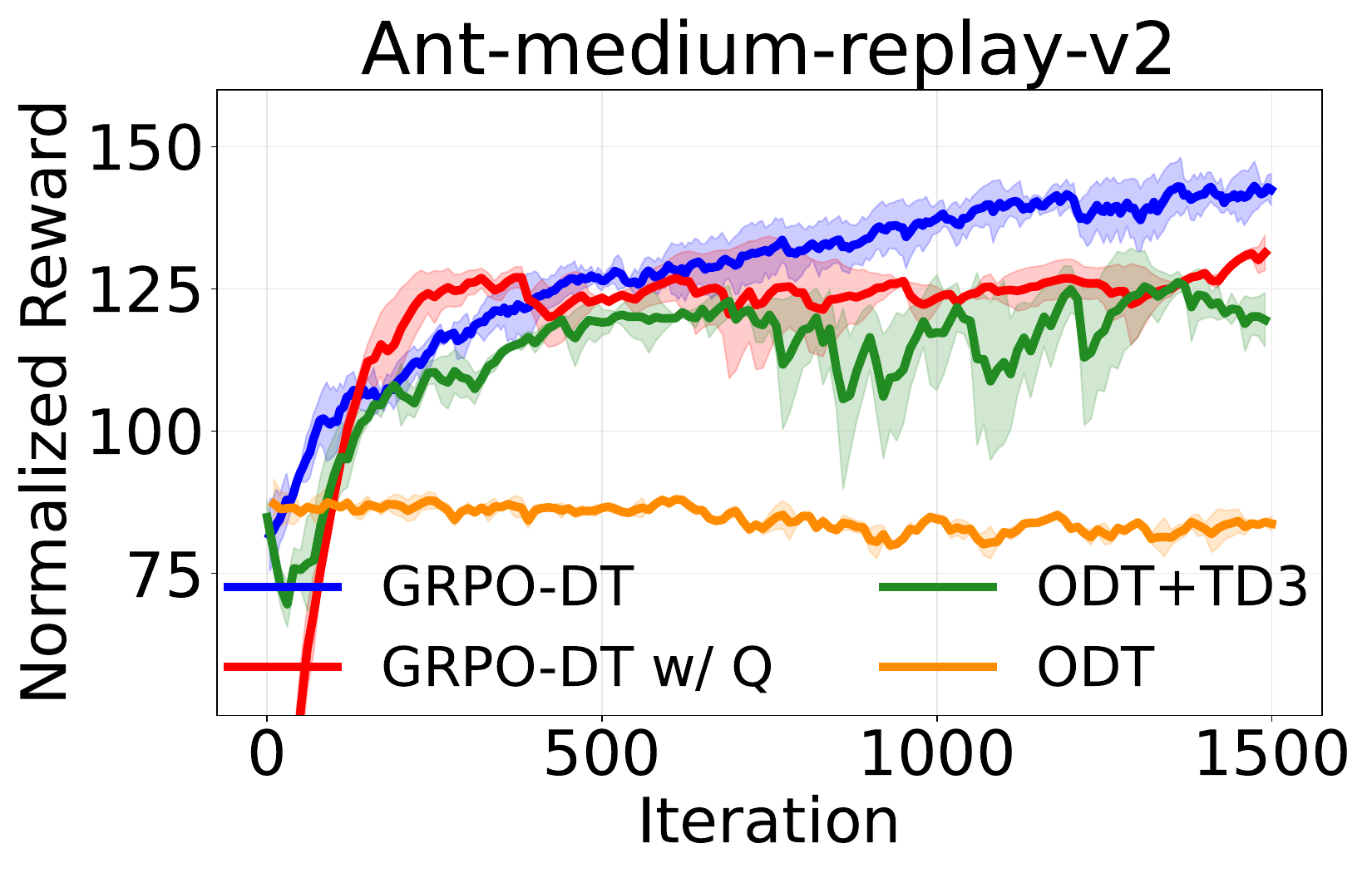}
\subcaption{Q-guided \ouralgtext}
\label{subfig:QGRPO}
\end{minipage}\hfill
\begin{minipage}{0.24\textwidth}
\centering
\includegraphics[width=\textwidth]{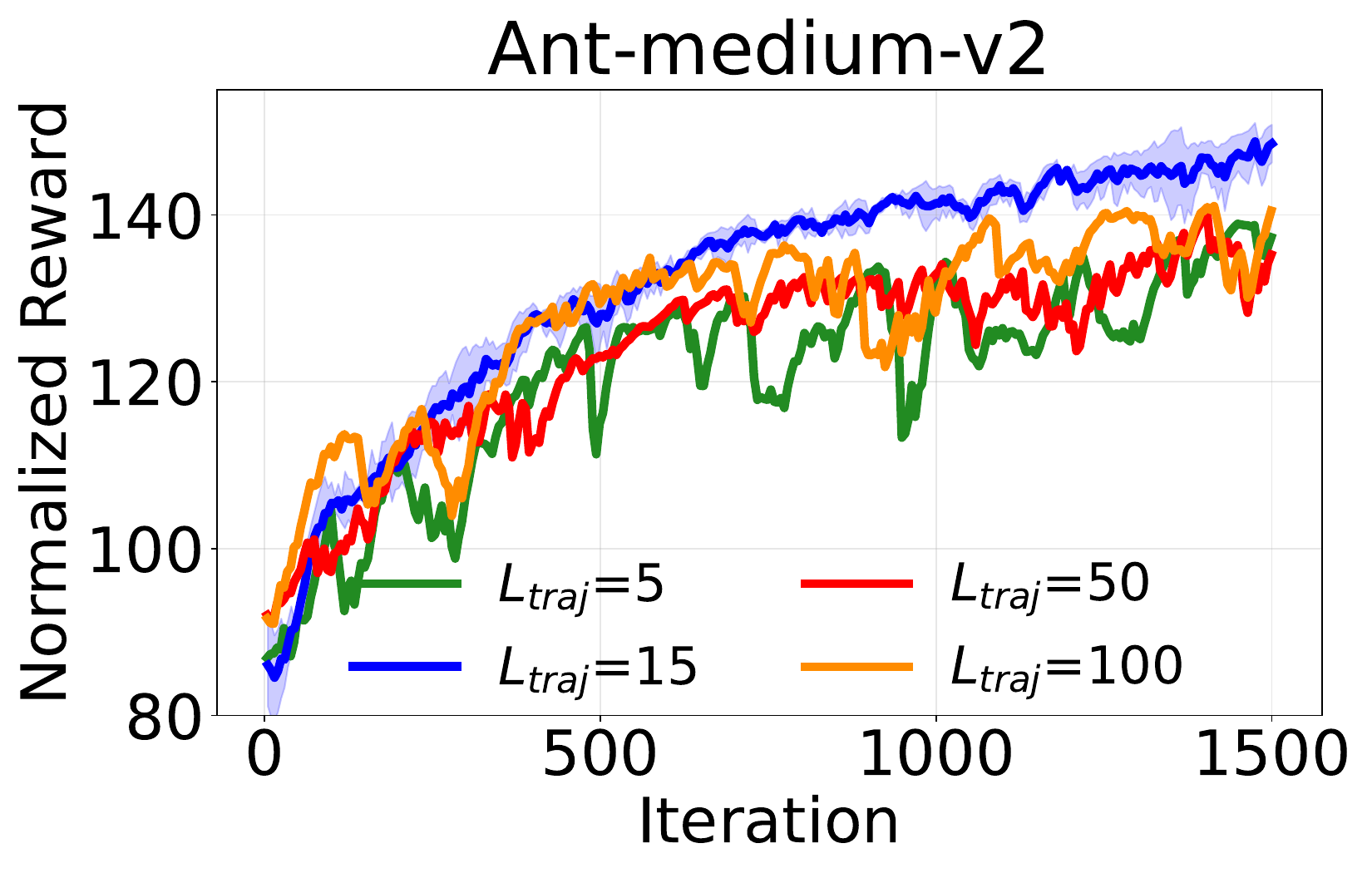}
\subcaption{Ablation on $L_\traj$}
\label{subfig:subtrajlen}
\end{minipage}\hfill
\begin{minipage}{0.24\textwidth}
\centering
\includegraphics[width=\textwidth]{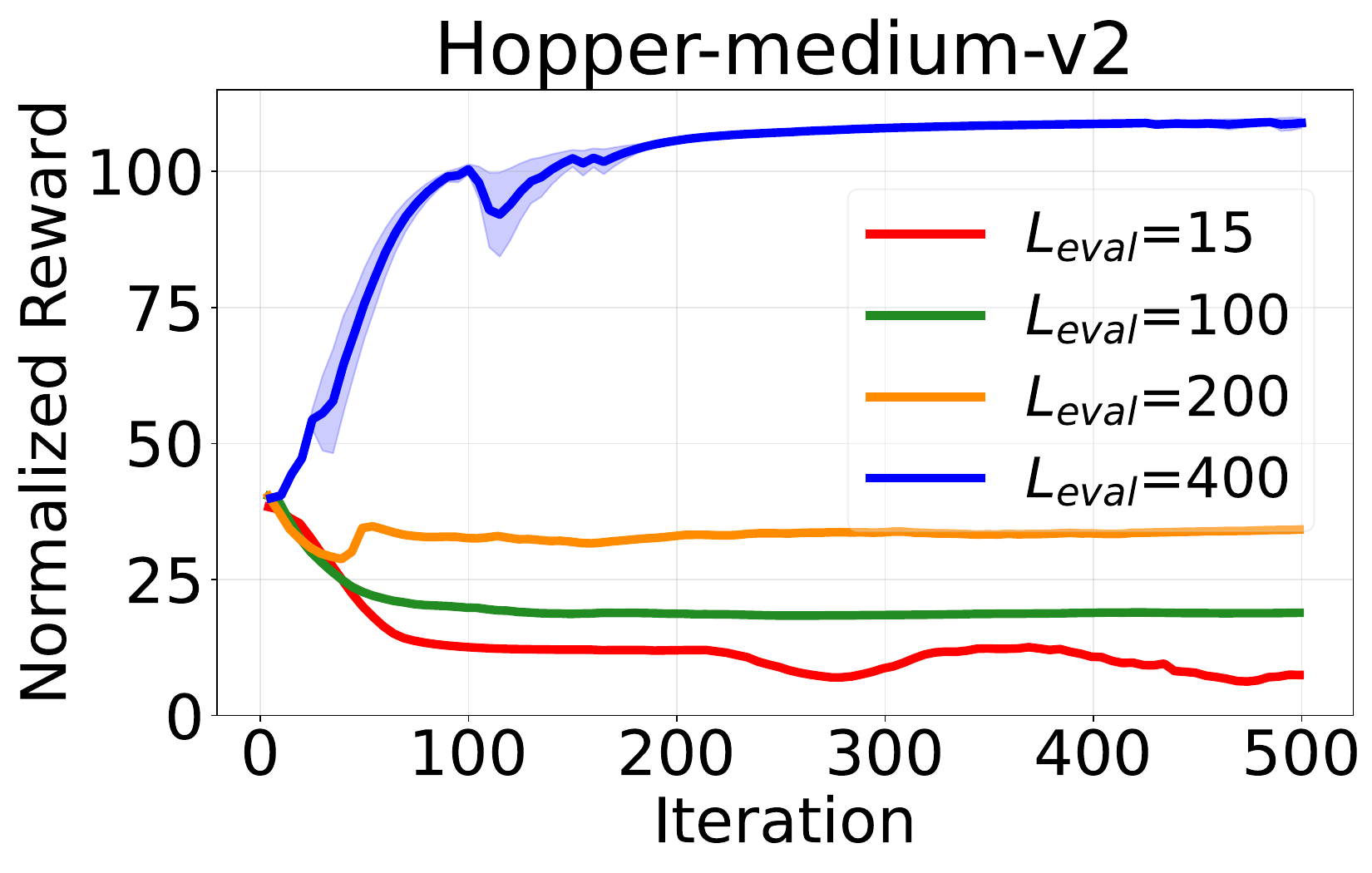}
\subcaption{Ablation on $L_\eval$}
\label{subfig:evaltrajlen}
\end{minipage}
\caption{
Additional analyses and ablations.
(a) Performance of \ouralgtext across different Decision Transformer models, including the standard DT and Reinformer, demonstrating that \ouralgtext generalizes beyond the standard DT.
(b) Comparison between reset-based \ouralgtext and Q-guided \ouralgtext when environment resetting is infeasible; Q-guided \ouralgtext maintains strong performance even without resetting.
(c) Ablation on sub-trajectory length $L_{\traj}$: both overly short and overly long sub-trajectories degrade performance, highlighting the importance of an appropriate sub-trajectory length.
(d) Ablation on sub-trajectory evaluation length $L_{\eval}$: insufficient evaluation steps lead to unstable training or model collapse, while larger $L_{\eval}$ yields more reliable performance.
} 
\label{fig:ablation_relabeling}
\end{figure}

\paragraph{Online finetuning beyond DTs.}
\label{sec:reinformer}
To assess the applicability of our \ouralgtext\ beyond the standard Decision Transformer, we evaluate it on the \textit{Reinformer} model \citep{zhuang2024reinformer}.
Reinformer integrates return maximization into supervised sequence modeling by leveraging expectile regression to predict higher RTGs, which then guide action selection.
As shown in \cref{subfig:architectures}, our \ouralgtext also performs effectively on Reinformer, suggesting that our proposed \ouralgtext generalize beyond standard DTs.

\paragraph{Q-guided \ouralgtext without resetting.}
In scenarios where environment resetting is infeasible, we train an auxiliary Q-function using \tdthree \citep{fujimoto2018addressing} and apply \ouralgtext\ with Q-guided advantages (see \cref{sec:qgrpo} for details).
This variant also reduces sample complexity, as the sub-trajectory generation and evaluation step (line~\ref{line:reset} in \cref{alg:grpowdt}) is no longer required.
As shown in \cref{subfig:QGRPO}, the Q-guided version of \ouralgtext\ maintains strong performance even without environment resetting.

\paragraph{Ablation on sub-trajectory length.} 
In our method, each sub-trajectory serves as the fundamental unit for advantage computation and credit assignment, making its length $L_{\traj}$ a critical hyperparameter.
Empirical results in \cref{subfig:subtrajlen} show that increasing the sub-trajectory length destabilizes training and degrades performance.
Conversely, very short sub-trajectories can also lead to suboptimal results.
This is likely because short segments sampled from similar state distributions are overly homogeneous, providing limited diversity and weaker learning signals for RL finetuning.

\paragraph{Ablation on sub-trajectory evaluation length.}
For each sub-trajectory, we extend the rollout by $L_\eval$ additional evaluation steps to compute rewards.
In our experiments, $L_\eval$ ranges from 30 to 400 depending on the environment (see \cref{tab:domain-specific}).
We conduct an ablation study on the evaluation length $L_\eval$ to examine its impact.
As shown in \cref{subfig:evaltrajlen}, increasing $L_\eval$ provides more reliable estimates of sub-trajectory quality, leading to consistent improvements in model performance.

\section{Related Work}
\label{sec5:relatedwork}
\paragraph{Transformers for RL.} 
With the transformer architecture \citep{vaswani2017attention} becoming the dominant paradigm in deep learning---most notably in language \citep{radford2018improving,brown2020language} and vision \citep{dosovitskiy2021an}---a growing body of work has explored transformer-based approaches for reinforcement learning. In this paradigm, exemplified by Decision Transformers \citep{chen2021decision}, RL is formulated as a sequence modeling problem, where models condition on past states, actions, and returns-to-go to autoregressively predict future actions. Leveraging the strong sequence modeling capability of transformers, DT-style methods have demonstrated competitive or state-of-the-art performance across a range of benchmarks \citep{janner2021offline,wang2022bootstrapped,yamagata2023q,zhuang2024reinformer}. While initially developed for offline RL, several recent works have extended these methods to the online setting---often referred to as offline-to-online RL \citep{lee2022offline,yu2023actor,nakamoto2023cal}. 
However, existing online finetuning approaches for Decision Transformers either rely purely on supervised sequence-modeling objectives \citep{zheng2022online,xie2023future}, or continue to prioritize the supervised loss by assigning only a small weight to reinforcement learning gradients \citep{yan2024reinforcement}.
In contrast, our work complements this line of research by presenting the first approach to online finetuning of Decision Transformers using \emph{pure reinforcement learning gradients}, enabling principled policy optimization without reliance on supervised objectives.
\looseness=-1

\paragraph{RL for transformers.} 
Reinforcement learning has also emerged as a powerful approach for aligning and enhancing large language models \citep{ouyang2022training,bai2022constitutional,lee2023rlaif} as well as multimodal models \citep{liu2025visual,shen2025vlm}. A broad spectrum of algorithms has been explored, ranging from online policy gradient methods such as PPO and its variants \citep{ouyang2022training,luong2024reft,kazemnejad2025vineppo}, to offline reinforcement learning methods based on IQL and its variants \citep{snell2022offline,qi2024verifierq}, as well as preference-based policy optimization approaches such as DPO and its extensions \citep{rafailov2023direct,ethayarajh2024kto}. More recently, substantial attention has been devoted to GRPO and its variants \citep{shao2024deepseekmath,guo2025deepseek,liu2025understanding,yu2025dapo}, which have been successfully applied to large language and multimodal models to improve reasoning performance. In this work, we extend GRPO and PPO to the online finetuning of Decision Transformers in classical reinforcement learning environments. Along the way, we also introduce a sub-trajectory variant of GRPO that significantly improves credit assignment over standard GRPO, and which may be of independent interest beyond the settings considered in this paper.

\paragraph{Additional related work.}
During the preparation of this paper, we became aware of a concurrent work on sequence-level importance ratios by \citet{zheng2025group}, which appeared online in July 2025. While our approach shares the same core idea, our sequence-level importance ratio technique was developed independently; an initial version of our work was made publicly available in September 2025. Our method differs from \citet{zheng2025group} in a key design choice. Specifically, while \citet{zheng2025group} propose using a geometric mean over sequence-level importance ratios, we remove this geometric normalization when performing sub-trajectory-level optimization. Empirically, we find that this modification yields superior performance when combined with relatively short rollout lengths (\cref{subfig:geometricmean}). We hypothesize that this improvement arises because learning without geometric normalization (i) more rapidly suppresses outdated sub-trajectories via clipping, and (ii) enables more aggressive updates for approximately on-policy sub-trajectories.
\looseness=-1

Our active state sampling technique is inspired by classical active learning \citep{settles2009active}, where uncertainty-based methods are known to exponentially improve sample complexity compared to standard supervised learning \citep{hanneke2014theory,puchkin2021exponential,zhu2022active,zhu2022efficient}, and have demonstrated strong empirical performance with deep neural networks \citep{Ash2020Deep,saran2023streaming,zhang2024labelbench}. More recently, the core principles of active learning have been applied to reinforcement learning \citep{yin2023sample}, large language models \citep{bhatt2024experimental, wang2025beyond}, and multimodal models \citep{zhang2025towards}.
\looseness=-1

\section{Conclusion}
\label{sec6:conclusion}

In this work, we conducted a systematic study of online finetuning Decision Transformers using \emph{pure reinforcement learning gradients}. We identify hindsight return relabeling as a key obstacle that hinders the effective application of importance sampling-based policy gradient methods to online DT finetuning. Building on this insight, we propose a principled solution by adapting GRPO to the Decision Transformer framework. Our approach integrates sub-trajectory optimization for improved credit assignment, sequence-level importance ratios for enhanced stability and efficiency, and active state sampling for better exploration, collectively enabling pure-RL online finetuning of pretrained DTs. We further demonstrate that PPO can also be adapted to online DT finetuning. Extensive experiments show that our methods outperform existing online DT baselines and achieve new state-of-the-art performance across multiple benchmarks. Overall, our findings highlight the viability and effectiveness of pure RL optimization for Decision Transformers and open new avenues for scaling DT-based agents through RL-driven finetuning.

\newpage
\bibliographystyle{plainnat}
\bibliography{refs}

\newpage
\appendix
\section{Environment and Dataset Details}
\label{sec:envs&datasets}

We consider three continuous control and manipulation environments from D4RL \citep{fu2020d4rl}: \textit{MuJoCo}, \textit{Adroit}, and \textit{AntMaze}.
In total, we conduct experiments on 8 different tasks spanning 17 datasets with varying offline data quality.
We provide detailed descriptions of each environment, task, and dataset below.

\paragraph{MuJoCo environments.}

\begin{table}[t]
\caption{MuJoCo dataset size (in terms of total \#steps) and normalized final reward statistics.}
\label{tab:gymdataset}
\centering
\small
\begin{tabularx}{0.8\textwidth}{l*{2}{>{\centering\arraybackslash}X}}
\toprule
  Dataset & Size &Normalized Reward \\
\midrule
  \mbox{Hopper-medium-v2} &\num{999906} & 44.32\,{\scriptsize $\pm$\,12.27} \\
  \mbox{Hopper-medium-replay-v2} &\num{402000} & 14.98\,{\scriptsize $\pm$\,16.32}\\
  \mbox{Hopper-random-v2} & \num{999906} & 1.19\,{\scriptsize $\pm$\,1.16}\\
  \mbox{Walker2d-medium-v2} & \num{999995} & 62.09\,{\scriptsize $\pm$\,23.83}\\
  \mbox{Walker2d-medium-replay-v2} &\num{302000} & 14.84\,{\scriptsize $\pm$\,19.48}\\
  \mbox{Walker2d-random-v2}  &\num{999997} & 0.01\,{\scriptsize $\pm$\,0.09}\\
  \mbox{Ant-medium-v2} &\num{999946} & 80.30\,{\scriptsize $\pm$\,35.82}\\
  \mbox{Ant-medium-replay-v2} &\num{302000} & 30.95\,{\scriptsize $\pm$\,31.66}\\
  \mbox{Ant-random-v2} & \num{999930} & 6.36\,{\scriptsize $\pm$\,10.07}\\
\bottomrule
\end{tabularx}
\end{table}

The first environment is \textbf{MuJoCo} \citep{todorov2012mujoco}, including locomotion tasks 
\textit{Hopper}, \textit{Walker2d}, and \textit{Ant} with dense reward signals.
We evaluate these tasks using the \textit{medium}, \textit{medium-replay}, and \textit{random} datasets, where the \textit{medium} dataset contains trajectories generated by a policy early-stopped at medium-level performance, the \textit{medium-reply} dataset contains trajectories sampled in the training process of the medium policy, and the \textit{random} contains trajectories collected by a random policy.
Dataset size and normalized reward statistics of each offline dataset is provided in \cref{tab:gymdataset}.
We discuss each task below.

\begin{itemize}
    \item \textbf{Hopper.} Hopper is a single-legged locomotion task where the agent controls three joints to make the robot hop forward while maintaining stability. The action space is a 3-dimensional continuous vector, corresponding to torques applied at the joints, each bounded within $[-1,1]$. The observation space has 11 dimensions, consisting of positional and velocity information. At each timestep, the reward is a combination of a healthy reward, a forward progress reward, and a control cost penalty proportional to the squared magnitude of the action. 
    \item \textbf{Walker2d.} Walker2d is a 2-dimensional bipedal walking robot task where the agent controls six joints to make the robot walk forward steadily. The action space is a 6-dimensional continuous vector, corresponding to torques applied at hinge joints, each bounded within $[-1, 1]$. The observation space has 17 dimensions, consisting of positional and velocity information. At each timestep, the reward is a combination of a healthy reward, a forward progress reward, and a control cost penalty proportional to the squared magnitude of the action. 
    \looseness=-1
    \item \textbf{Ant.} Ant is a 3-dimensional locomotion task where the agent controls an 8-joint quadruped to move forward while maintaining balance. The action space is a 8-dimensional continuous vector, corresponding to torques applied at hinge joints, each bounded within $[-1, 1]$. The observation space has 105 dimensions 
    consisting of positional, velocity, and external contact force information.
    At each timestep, the reward is a combination of a healthy reward, a forward progress reward, a control cost penalty proportional to the squared magnitude of the action, and an external contact force penalty proportional to the squared magnitude of contact force. 

\end{itemize}

\paragraph{Adriot environment.}
The second environment is \textbf{Adroit}, including challenging manipulation tasks \textit{Door}, \textit{Hammer}, and \textit{Pen}.
We evaluate these tasks using the \textit{human} and \textit{cloned} datasets, where the \textit{human} dataset contains a small amount of human demonstrations, and the \textit{cloned} dataset contains the mixture of trajectories collected from a behavior cloning policy and human demonstrations. Dataset size and normalized reward statistics of each offline dataset is provided in \cref{tab:adroitdataset}.
We discuss each task below.

\begin{table}[t]
\caption{Adroit dataset size (in terms of total \#steps) and normalized final reward statistics.}
\label{tab:adroitdataset}
\centering
\small
\begin{tabularx}{0.8\textwidth}{l*{2}{>{\centering\arraybackslash}X}}
\toprule
Dataset & Size & \mbox{Normalized Reward} \\
\midrule
Pen-cloned-v1 & \num{499886} &108.63\,{\scriptsize $\pm$\,122.43}\\
Pen-human-v1 & \num{4800} & 202.69\,{\scriptsize $\pm$\,154.48}\\
Hammer-cloned-v1 & \num{999872} & 8.11\,{\scriptsize $\pm$\,23.35}\\
Hammer-human-v1 & \num{10948} & 23.80\,{\scriptsize $\pm$\,33.86}\\
Door-cloned-v1 & \num{999939} & 12.29\,{\scriptsize $\pm$\,18.35}\\
Door-human-v1 & \num{6504} & 28.35\,{\scriptsize $\pm$\,13.88}\\
\bottomrule
\end{tabularx}
\end{table}

\begin{itemize}
    \item \textbf{Door.} The Door task involves a robotic hand-arm system that learns to unlatch and open a door.
The action space is a 28-dimensional continuous vector representing joint angular positions (scaled to $[-1, 1]$).
The 39-dimensional observation space contains information about the angular position of the finger joints, the pose of the palm of the hand, as well as state of the latch and door.
The dense reward combines distance, hinge-alignment, and velocity penalties with positive rewards for door hinge displacement.
  
   \item \textbf{Hammer.} The Hammer task involves a robotic hand-arm system to pick up a hammer and drive a nail into a board. The action space is a 26-dimensional continuous vector representing joint angular positions (scaled to $[-1, 1]$).
   The 46-dimensional observation space contains information about the angular position of the finger joints, the pose of the palm of the hand, the pose of the hammer and nail, and external forces on the nail.
   The dense reward combines distance and velocity penalties with positive rewards for lifting the hammer and driving the nail.
    
   \item \textbf{Pen.} The Pen task involves a robotic hand-arm system to manipulate a pen into a target orientation. The action space is a 24-dimensional continuous vector representing joint angular positions (scaled to $[-1, 1]$).
   The 45-dimensional observation space contains information about the angular position of the finger joints, the pose of the palm of the hand, as well as the pose of the real pen and target goal.
   The dense reward combines distance and orientation penalties and a penalty for dropping the pen, with bonuses for precise alignment and stable control.
\end{itemize}

\paragraph{AntMaze environment.}
The third environment is \textbf{AntMaze} \citep{fu2020d4rl}, including navigation tasks in mazes with sparse goal-reaching rewards.
We evaluate on the \textit{umaze} and \textit{umaze-diverse} tasks, where the former has fixed starting and ending points and the later has random starting/ending points.
The umaze environment in AntMaze places an ant quadruped in a U-shaped maze. The action space is a 8-dimensional continuous vector, corresponding to torques applied at hinge joints, each bounded within $[-1, 1]$.
The observation space is a goal-aware dictionary, consisting of a 27-dimensional “observation” vector (positions and velocities of the Ant body parts), a 2-dimensional desired goal vector (coordinates of the desired position), and a 2-dimensional achieved goal vector (coordinates of the current position).
The reward is sparse, granting a value of 1 when the ant reaches the target position and 0 otherwise.
Dataset size and normalized reward statistics of both offline dataset is provided in \cref{tab:antmazedataset}.\footnote{Following \odttdthree's \citep{yan2024reinforcement} practice, we remove all 1-step trajectories in the dataset.}

\begin{table}[t]
\caption{AntMaze dataset size (in terms of total \#steps) and normalized final reward statistics.}
\label{tab:antmazedataset}
\centering
\small
\begin{tabularx}{0.8\textwidth}{l*{2}{>{\centering\arraybackslash}X}}
\toprule
Dataset & Size & Normalized Reward \\
\midrule
Antmaze-umaze-v2 & \num{998573} & 86.14\,{\scriptsize $\pm$\,34.55}\\
Antmaze-umaze-diverse-v2 & \num{999000} & 3.48\,{\scriptsize $\pm$\,18.32}\\
\bottomrule
\end{tabularx}
\end{table}

\section{Experimental Details}
\subsection{Hyperparameters}
\label{sec:hyperparams}

In this section, we describe the hyperparameters used in our experiments. For the \odttdthree and \odt baselines, we use the codebase and default hyperparameters provided by \citet{yan2024reinforcement}. For the IQL baseline, we largely follow the implementation of \citet{yan2024reinforcement}, but set the number of pretraining steps to match those used by our methods and other baselines to ensure a fair comparison.

\cref{tab:common_params} summarizes the common architectural and training hyperparameters used by our algorithms across all environments, which largely follow the practices of \citet{yan2024reinforcement}. Following \citet{zheng2022online} and \citet{yan2024reinforcement}, we do not use positional embeddings. Additional algorithm-specific details are provided below.

For \ouralgtext, we collect 1 full trajectory per iteration in MuJoCo and AntMaze, and 5 full trajectories per iteration in Adroit. The full-trajectory replay buffer stores up to 32 trajectories. During resetting, we sample 16 trajectories from the buffer, select 4 reset points per trajectory, and construct groups of size 8, resulting in 512 sub-trajectories per iteration. The sub-trajectory buffer stores up to 2048 sub-trajectories. When forming groups of sub-trajectories, we discard those whose raw rewards are within an additive margin $\Delta_r$ of the group-average reward, in order to provide stronger optimization signals. Environment-specific hyperparameters for \ouralgtext are reported in \cref{tab:domain-specific}.

For Q-guided \ouralgtext, we conduct experiments on Ant-medium-replay-v2 (\cref{subfig:QGRPO}) and largely follows the practice of \citet{yan2024reinforcement} for training the Q-functions and our \ouralgtext for training the policy network. To improve training stabilities, we change the policy learning rate to $\lrmath_{\policymath} = 1 \times 10^{-5}$ and the initial entropy dual variable to $\kappa_\initialmath = 1 \times 10^{-2}$. 

For \ourppotext, we collect 8 full trajectories per iteration in MuJoCo and AntMaze, and either 8 or 16 full trajectories per iteration in Adroit (see \cref{tab:ppo-domain-specific}). The full-trajectory replay buffer stores 4 times the number of trajectories collected per iteration. Following the practice of \odttdthree, we apply LayerNorm \citep{ba2016layer} to the value network in Adroit and AntMaze to improve training stability. Environment-specific hyperparameters for \ourppotext are reported in \cref{tab:ppo-domain-specific}.

\begin{table}[t]
\caption{Common architecture and training hyperparameters used across all environments.
The policy network is parameterized as a Transformer, while the value network is parameterized as an MLP.}
\label{tab:common_params}
\centering
\small
\begin{tabularx}{\textwidth}{l*{1}{>{\centering\arraybackslash}X}}
\toprule
  Hyperparameters for policy network (Transformer) & Value \\
\midrule
  Number of layers &4\\
  Number of attention heads &4\\
  Embedding dimension &512\\
  Activation function &SiLU \citep{elfwing2018sigmoid}\\
  Optimizer &LAMB \citep{you2019large}\\
  Dropout &0.1 in pretraining, disabled in finetuning\\
  Gradient norm clip & $0.5$\\
  Weight decay & $1 \times 10^{-4}$\\
  KL coefficient $\beta$ & $1 \times 10^{-3}$ \\
  Target entropy $\rho$ & $-\mathsf{dim}(\mathcal{A}$) \\
  \midrule
  Hyperparameters for value network (MLP) & Value \\
\midrule
  Number of layers &2\\
  Embedding dimension &256 for MuJoCo, 512 for others\\
  Activation function &SiLU \citep{elfwing2018sigmoid}\\
  Optimizer &AdamW \citep{loshchilov2017decoupled}\\
\bottomrule
\end{tabularx}
\end{table}

\begin{table}[t]
\caption{
Environment-specific hyperparameters for \ouralgtext.
$n_\batchmath$ denotes the training batch size.
$c_\offlinemath$ and $c_{\onlinemath}$ denote the context lengths used in the offline and online stages, respectively.
$g_\onlinemath$ denotes the initial return-to-go for online exploration.
$L_{\traj}$ denotes the sub-trajectory rollout length, and $L_{\eval}$ denotes the number of additional evaluation steps.
$\gamma$ is the discount factor used to compute sub-trajectory rewards.
$\epsilon$ denotes the GRPO clipping hyperparameter, and $\Delta_r$ denotes the additive margin used for sub-trajectory filtering.
$\lrmath_\policymath$ denotes the learning rate for the policy network, and $\kappa_\initialmath$ denotes the initial value of the entropy dual variable for online finetuning, which is adaptively updated during training.
}
\label{tab:domain-specific}
\centering
\scriptsize
\begin{tabularx}{\textwidth}{l *{11}{>{\centering\arraybackslash}X}}
\toprule
Environments 
& $n_\batchmath$
& $c_\offlinemath$ 
& $c_\onlinemath$ 
& $g_\onlinemath$
& $L_{\traj}$ 
& $L_{\eval}$
& $\gamma$
& $\epsilon$
& $\Delta_r$
& $\lrmath_\policymath$
& $\kappa_{\initialmath}$ \\
\midrule
\mbox{Ho-M/MR-v2} &256 &20&1&7200&15&400&0.995&0.2&2.0&5e-5&0.20\\
\mbox{Ho-R-v2}    &256 &20&1&7200&15&400&0.995&0.2&2.0&5e-5&0.20\\
\mbox{Wa-M/MR-v2} &256 &20&1&10000&15&400&0.995&0.3&2.0&5e-5&0.04\\
Wa-R-v2           &256 &20&1&10000&15&400&0.995&0.3&2.0&5e-5&0.20\\
\mbox{An-M/MR-v2} &256 &20&1&12000&15&200&0.995&0.3&2.0&5e-5&0.04\\
An-R-v2           &256 &20&1&12000&15&200&0.995&0.3&2.0&5e-5&0.20\\
\midrule
D-C-v1 &512&5&1&3000&10&100&0.99&0.3&0.5&3e-5&0.10\\
D-H-v1 &512&5&1&3000&10&100&0.99&0.3&0.4&3e-5&0.04\\
P-C-v1 &512&5&1&6000&3&30&0.99&0.3&0&3e-5&0.02\\
P-H-v1 &512&5&1&6000&3&30&0.99&0.3&0&3e-5&0.02\\
H-C-v1 &512&5&5&4000&10&100&0.99&0.3&0&3e-5&0.05\\
H-H-v1 &512&5&5&4000&10&100&0.99&0.3&0.8&3e-5&0.05\\
\midrule
U-v2  &256&5&1&2&10&200&1.0&0.2&0&5e-5&0.05\\
UD-v2 &256&1&5&2&10&200&1.0&0.2&0&5e-5&0.05\\
\bottomrule
\end{tabularx}
\end{table}

\begin{table}[t]
\caption{
Environment-specific hyperparameters for \ourppotext.
$n_\batchmath$ denotes the training batch size.
$c_\offlinemath$ and $c_{\onlinemath}$ denote the context lengths used in the offline and online stages, respectively.
$g_\onlinemath$ denotes the initial return-to-go for online exploration.
$K_{\mathsf{PPO}}$ denotes the number of online full trajectories rollouts per iteration.
$\lrmath_\policymath$ and $\lrmath_\valuemath$ denote the learning rates for the policy and value networks, respectively.
$\kappa_\initialmath$ denotes the initial value of the entropy dual variable for online finetuning, which is adaptively updated during training.
Across all environments, we set the PPO clipping hyperparameter $\epsilon = 0.2$ and the GAE hyperparameters $(\gamma, \lambda) = (0.99, 0.95)$.
}
\label{tab:ppo-domain-specific}
\centering
\scriptsize
\begin{tabularx}{\textwidth}{l *{8}{>{\centering\arraybackslash}X}}
\toprule
Environments 
& $n_\batchmath$ 
& $c_\offlinemath$ 
& $c_\onlinemath$ 
& $g_{\onlinemath}$ 
& $K_{\mathsf{PPO}}$
& $\lrmath_\policymath$ 
& $\lrmath_\valuemath$ 
& $\kappa_{\initialmath}$ \\
\midrule
\mbox{Ho-M/MR-v2} &256 &20&1&7200 &8  &5e-5&1e-3&0.02\\
\mbox{Ho-R-v2}    &256 &20&1&7200 &8  &5e-5&1e-3&0.04\\
\mbox{Wa-M/MR-v2} &256 &20&1&10000&8  &5e-5&1e-3&0.02\\
Wa-R-v2           &256 &20&1&10000&8  &5e-5&1e-3&0.20\\
\mbox{An-M/MR-v2} &256 &20&1&12000&8  &5e-5&1e-3&0.02\\
An-R-v2           &256 &20&1&12000&8  &5e-5&1e-3&0.02\\
\midrule
D-C-v1 &512&5&1&3000 &16 &3e-5&2e-4&0.002\\
D-H-v1 &512&5&1&3000 &16 &3e-5&2e-4&0.002\\
P-C-v1 &512&5&1&6000 &8  &3e-5&2e-4&0.04\\
P-H-v1 &512&5&1&6000 &8  &3e-5&2e-4&0.04\\
H-C-v1 &512&5&5&4000 &16 &3e-5&2e-4&0.005\\
H-H-v1 &512&5&5&4000 &16 &3e-5&2e-4&0.005\\
\midrule
U-v2  &256&5&1&2 &8 &5e-5&1e-3&0.02\\
UD-v2 &256&1&5&2 &8 &5e-5&1e-3&0.02\\
\bottomrule
\end{tabularx}
\end{table}

\subsection{Implementation Details of \ourppotext}
\label{sec:ppo}

In this section, we present the implementation of our \ourppotext (\cref{alg:ppo}), which adapts the classical Proximal Policy Optimization (PPO, \citealp{schulman2017proximalpolicyoptimizationalgorithms}) to Decision Transformers \citep{chen2021decision}.

In each iteration, we sample $K_{\mathsf{PPO}}$ complete trajectories. For each trajectory $\tau$ and each time step $h$, we compute the advantage via generalized advantage estimation (GAE, \citealp{schulman2015high}):
\begin{equation} \wh {A}_{h} = \sum_{\ell=0}^{T-h-1} (\gamma\lambda)^\ell \Big(r_{{h+\ell}} + \gamma V_{\phi_t}(s_{{h+\ell+1}}) - V_{\phi_t}(s_{{h+\ell}})\Big), \label{eq:gae} \end{equation}
where $V_{\phi_t}$ denote the learned value function, $\gamma \in [0,1]$ denotes the discount factor and $\lambda \in [0,1]$ controls the bias-variance trade-off of GAE.
To update the policy network, we first randomly sample sub-trajectory $\tau^{\sub}_k = (s_{k_1}, a_{k_1}, g_{k_1}, \cdots, s_{k_L}, a_{k_L}, g_{k_L})$ of length $L = c_\train$ to form a batch $\cB_\sub$; following the practice in \citet{huang2022cleanrl}, we further normalize the advantages across the batch. We then update the parameter by maximizing the following objective: 
\begin{align}
    J_{\ppomath}(\theta) = &
    \frac{1}{\abs{\cB_\sub}\abs{\tau^\sub_k}} \sum_{\tau_k^\sub \in \cB_\sub}\sum_{i=1}^{\abs{\tau^\sub_k}} 
    \nonumber \\
    & \min\Bigg(
        \frac{\pi_\theta(a_{k_i} \mid s_{k_i}, g_{k_i})}
             {\pi_{\theta_{\old}}(a_{k_i} \mid s_{k_i}, g_{k_i})} \wh A_{k_i}, \,
        \mathsf{clip}\!\left(
            \frac{\pi_\theta(a_{k_i} \mid s_{k_i}, g_{k_i})}
             {\pi_{\theta_{\old}}(a_{k_i} \mid s_{k_i}, g_{k_i})}, 
            1 - \epsilon, 
            1 + \epsilon 
        \right) \wh A_{k_i}
    \!\Bigg) 
     - \beta D_{\mathsf{KL}}\!\left( \pi_\theta \, \| \, \pi_{\theta_\refmath} \right).
    \label{eq:ppo_policy}
\end{align}
Similar to \ouralgtext, 
we augment \cref{eq:ppo_policy} with an additional entropy term
$\kappa H(\pi_\theta(\tau_{k}^\sub \mid s_k, g_k))$,
while adaptively updating $\kappa$ during training to ensure that the entropy constraint
$H(\pi_\theta(\tau_{k}^\sub \mid s_k, g_k)) \ge \rho$
is approximately satisfied.

The value function is updated by minimizing the square loss with respect to value target $V_{\mathsf{target}(s_h)} = \wh A_{h} + V_{\phi_t}(s_h)$ on states present in the sampled sub-trajectories: 
\begin{equation} \mathcal{L}_{\mathsf{value}} = \mathbb{E}_{s_h}\!\left[(V_{\phi}(s_{h}) - V_{\text{target}}(s_{h}))^2\right]. \label{eq:ppocriticloss} \end{equation}

\begin{algorithm}[t]
\caption{Online Finetuning Decision Transformers with PPO (PPO-DT)}
\label{alg:ppo}
\renewcommand{\algorithmicrequire}{\textbf{Input:}}
\renewcommand{\algorithmicensure}{\textbf{Output:}}
\begin{algorithmic}[1]
\REQUIRE 
Pretrained policy $\pi_{\theta_1}$, value network $V_{\phi_1}$, full trajectory buffer $\mathcal{T}_{\mathsf{replay}}$, number of iterations $T$, initial return-to-go $g_{\onlinemath}$, number of trajectories collected per iteration $n_{\mathsf{PPO}}$, sub-trajectory length $\cltrain$.
\FOR{iteration $t = 1, \cdots, T$}
    \STATE Roll out $K_{\mathsf{PPO}}$ trajectories using the current policy 
    $\pi_{\theta_t}(\cdot \mid s_1, g_{\onlinemath})$, conditioned on initial (randomized) state $s_1$ and RTG $g_{\mathsf{online}}$;
    update $\mathcal{T}_{\mathsf{replay}}$. 
    \algcommentlight{Collect full trajectories; FIFO buffer update.}
    \STATE For each trajectory $\tau \in \mathcal{T}_{\mathsf{replay}}$, compute advantage for each step based on GAE (\cref{eq:gae}).
    \STATE Sample a batch $\mathcal{B}$ of full trajectories from $\mathcal{T}_{\mathsf{replay}}$ from distribution 
    $ p(\tau) = \frac{|\tau|}{\sum_{\tau \in \mathcal{T}_{\mathsf{replay}}} |\tau|}.$
    \STATE For each trajectory $\tau \in \cB$, randomly sample one sub-trajectory $\tau^\sub$ of length $L = c_\train$ to form the sub-trajectory batch $\cB_\sub$. Normalize the advantages across batch $\cB_\sub$.
    \STATE Finetune the current policy based on \cref{eq:ppo_policy} to get an updated $\pi_{\theta_{t+1}}$; finetune the value network based on \cref{eq:ppocriticloss} to get an updated $V_{\phi_{t+1}}$.
    
\ENDFOR
\ENSURE 
Online finetuned policy $\pi_{\theta_{T+1}}$. 
\end{algorithmic}
\end{algorithm}

\end{document}